\documentclass{siamltex}

\usepackage[text={5.125in,8.25in},centering]{geometry}
\usepackage{graphicx}
\usepackage{epsfig} 
\usepackage{mathptmx} 
\usepackage{times} 
\usepackage{amsmath} 
\usepackage{amssymb}  
\usepackage{dsfont}
\usepackage{verbatim}

\usepackage{color}
\usepackage[dvipsnames]{xcolor}

\definecolor{amethyst}{rgb}{1, 0, 1}
\definecolor{blue-violet}{rgb}{0.54, 0.17, 0.89}
\definecolor{brightturquoise}{rgb}{0.03, 0.91, 0.87}

\usepackage{caption}
\usepackage{subcaption}

\usepackage{algorithm,algpseudocode}
\newcommand*\Let[2]{\State #1 $\gets$ #2}
\algrenewcommand\algorithmicrequire{\textbf{Input:}}
\algrenewcommand\algorithmicensure{\textbf{Input:}}

\usepackage{pgfplots}\pgfplotsset{compat=1.4}

\usepackage{pgfplots}\pgfplotsset{compat=1.4}
\definecolor{softgray}{rgb}{0.92,0.92,0.95}
\definecolor{softblue}{rgb}{0.90,0.92,1.00}
\definecolor{lightgray}{rgb}{0.12,0.12,0.55}

\definecolor{theframe}{gray}{0.75}
\definecolor{theblue} {rgb}{0.02,0.04,0.48}
\definecolor{thegrey} {gray}{0.5}
\definecolor{theshade}{gray}{0.98}
\definecolor{thered}  {rgb}{0.00,0.00,0.00}
\definecolor{thegreen}{rgb}{0.3,0.3,0.3}

\usepackage{color}
\definecolor{softblue}{rgb}{0.90,0.92,1.00}
\usepackage{mdframed}
\usepackage{makecell}

\usepackage{hyperref}

\numberwithin{equation}{section}

\newtheorem{lemmer}{Lemma}[section]
\newtheorem{remark}[lemmer]{Remark}

\newcommand{\B}[1]{{\bf #1}}

\newcommand{\mB}[1]{{\mathbb{#1}}}

\DeclareMathOperator*{\argmin}{argmin}
\newcommand{\map}[3]{#1:#2\rightarrow #3}

\newcommand{\Nul}[1]{\mathrm{Nul}\left( #1\right)}

\newcommand{\F}{\mathcal{F}}

\newcommand{\blue}{\textcolor{black}}

\usepackage{fancyvrb}

\title{Generalized system identification with stable spline kernels}
 \author{%
    Aleksandr Y. Aravkin\thanks{Applied Mathematics Department, University of Washington, Seattle, WA,
      (saravkin@uw.edu).
     Research was supported by the Washington Research Foundation Fund for Innovation in Data-Intensive Discovery.}
    \and James V. Burke\thanks{Mathematics Dept., University
      of Washington, Seattle, WA 98195 (burke@math.washington.edu).
      Research is supported in part by the National Science Foundation under grant number DMS­-1514559.}
    \and Gianluigi Pillonetto
    \thanks{Control and Dynamic Systems Department of Information Engineering at the University of Padova, Padova, Italy (giapi@dei.unipd.it)}
    \hfill\today}
\begin{document}

\maketitle

\begin{abstract}
Regularized least-squares approaches have been 
successfully applied to linear system identification. 
Recent approaches use quadratic 
penalty terms on the unknown impulse response
defined by {\it stable spline kernels}, which control model space complexity by leveraging 
regularity and bounded-input bounded-output stability.
This  paper extends linear system identification to a wide class of nonsmooth stable spline estimators,
where regularization functionals and data misfits can be selected from a rich set of 
piecewise linear-quadratic (PLQ) penalties. This class includes 
the 1-norm, Huber, and Vapnik, in addition to the least-squares penalty. 

By representing penalties through their conjugates, the modeler can
specify {\it any} piecewise linear-quadratic penalty for misfit and regularizer, 
as well as inequality constraints on the response.  
\blue{The interior-point solver we implement (IPsolve) is locally quadratically convergent, with 
$O(\min(m,n)^2(m+n))$ arithmetic operations per iteration, where $n$ the number of unknown impulse response coefficients and $m$ the number of observed output measurements.
IPsolve is competitive with available alternatives for system identification. 
This is shown by a comparison with TFOCS, libSVM, and the FISTA algorithm. 
The code is open source\footnote{https://github.com/saravkin/IPsolve.}.}   

\blue{The impact of the approach for system identification is illustrated with numerical experiments }
featuring robust formulations for contaminated data, 
relaxation systems, nonnegativity and unimodality
constraints on the impulse response, \blue{and sparsity promoting regularization. 
Incorporating constraints yields particularly significant improvements. }
\end{abstract}

{\bf{Keywords}}: \small 
linear system identification; kernel-based
regularization; Gaussian processes; bias-variance; model order selection; robust statistics; 
sparse optimization; interior point methods

\section{Introduction}
\normalsize

System identification formalizes the process of inferring models from observations
and studying their properties. A {\it system} comprises multiple variable interactions  
to produce observable signals~\cite{Ljung}.
We focus on 
 {\it linear time invariant} system (LTI) identification, i.e. systems
where the response to a certain input signal does not depend on absolute time, and 
the output response to a linear combination of inputs is the linear combination 
of the responses to individual inputs. This class is computationally tractable, and has been successful
in a wide range of applications including
NMR spectroscopy, seismology, circuits, signal processing, control theory, 
biological processes, and many others. \blue{ Many techniques have been developed to identify LTI systems, 
 in state space and frequency domains (see eg.\cite{Ljung}).
We focus on on identifying impulse responses from input-output data in the time domain.} \\
Within the class of LTI systems, model selection and model space exploration is a key concept.
\begin{figure*}
  \begin{center}
   \begin{tabular}{cccc}
\hspace{-.4in}
 { \includegraphics[scale=0.3]{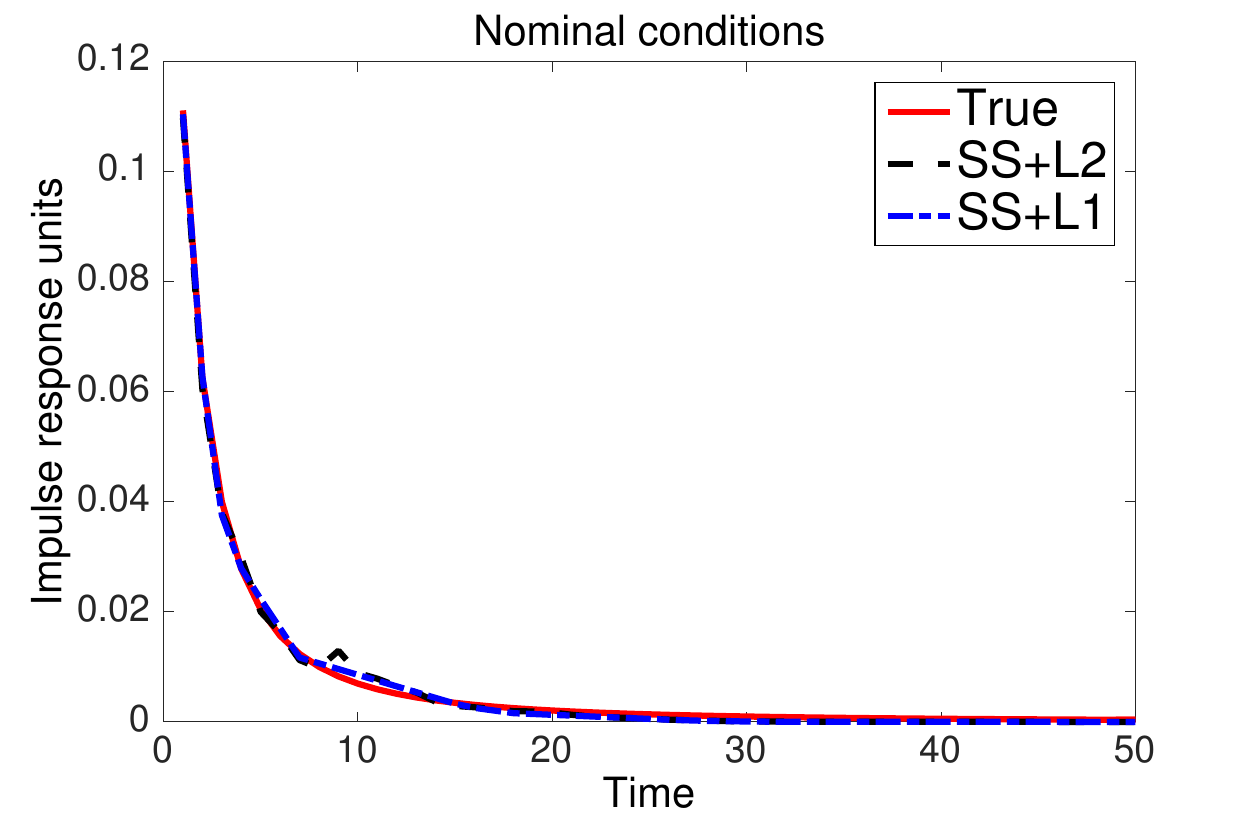}} 
 \hspace{.001in}
 { \includegraphics[scale=0.3]{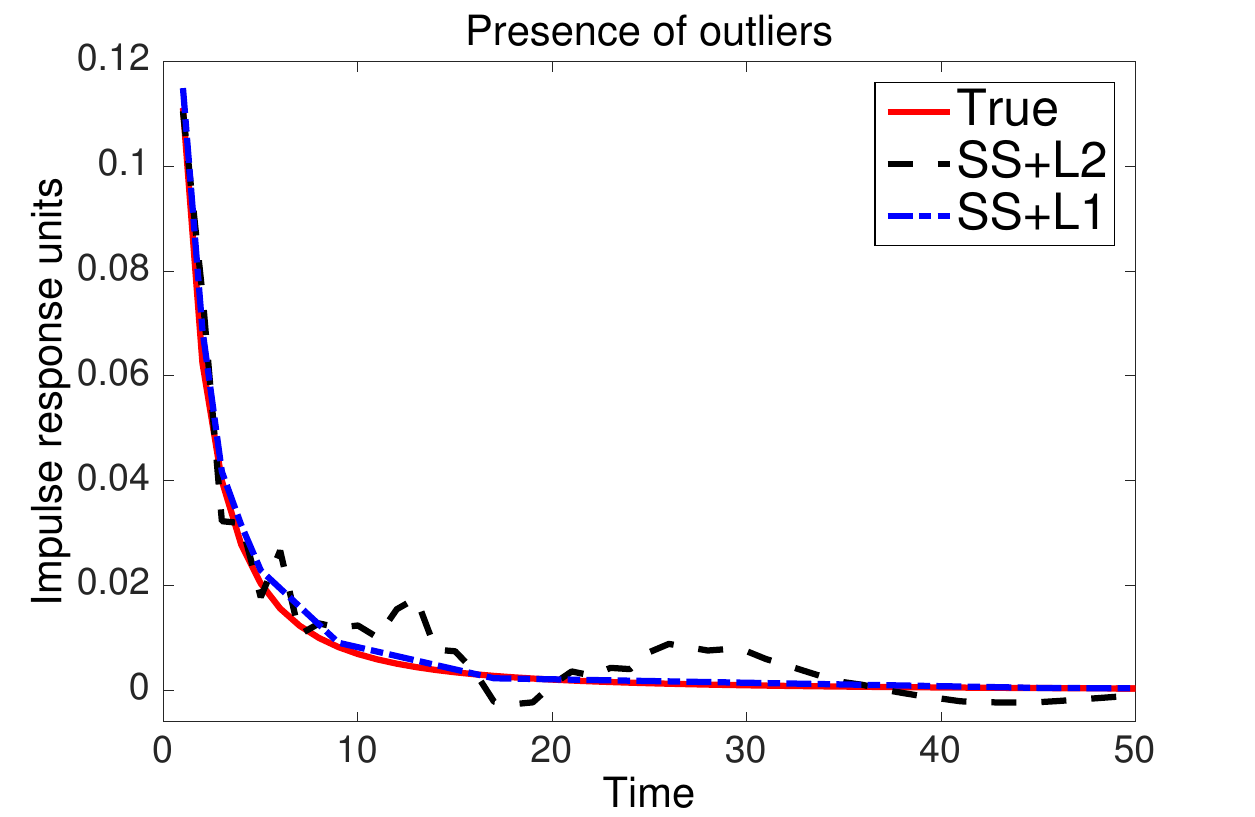}} \\
\hspace{-.4in}
 { \includegraphics[scale=0.3]{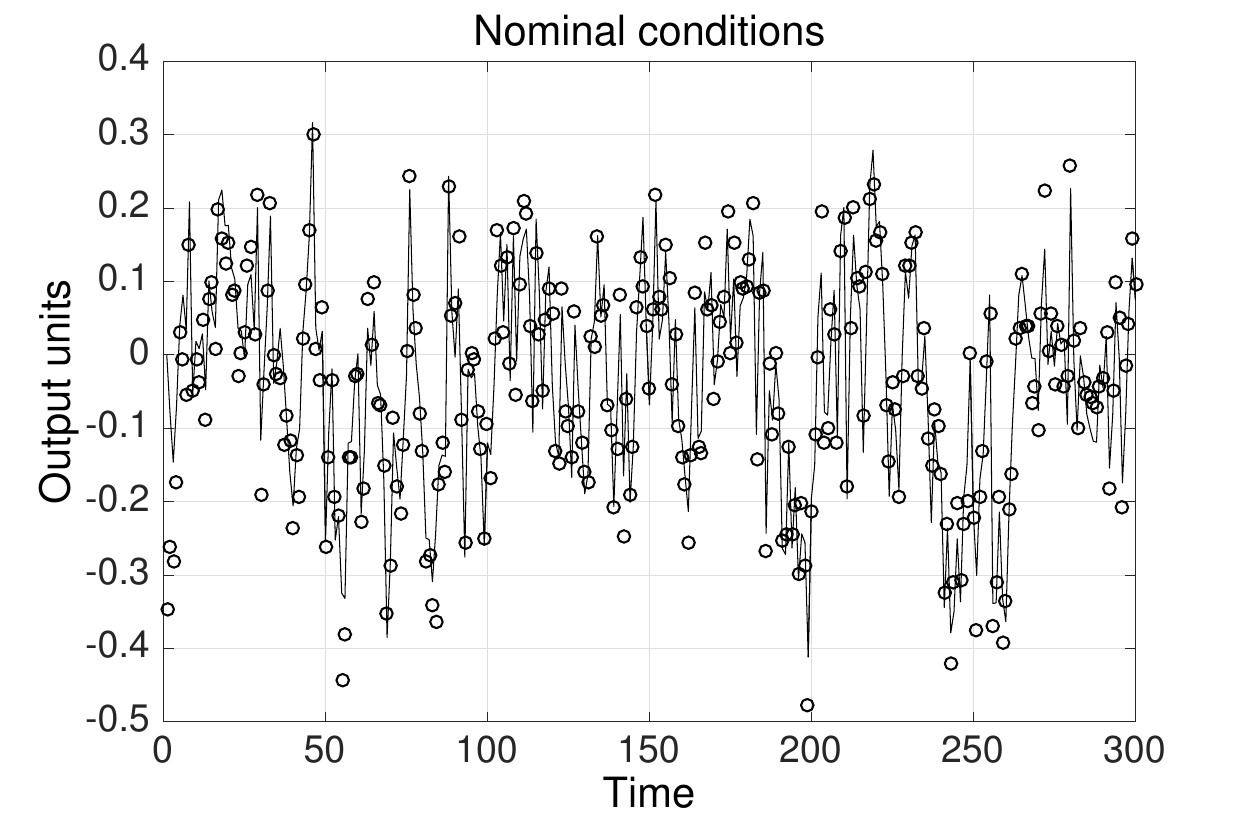}} 
\hspace{.001in}
{\includegraphics[scale=0.3]{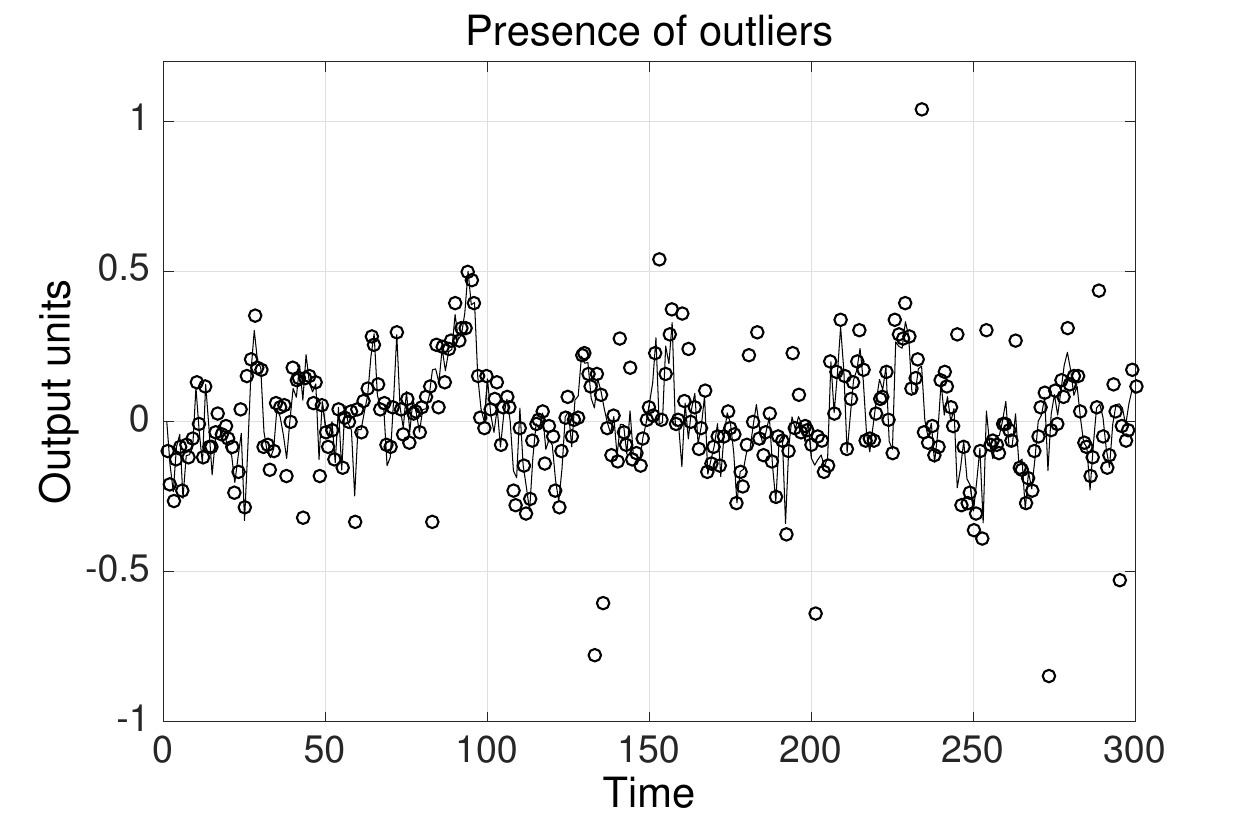}}
    \end{tabular}
 \caption{{\textit{Top:}} true impulse response (red solid) and 
estimates obtained under nominal conditions (left) and in presence of outliers (right) using the stable spline estimator (Section~\ref{SS+L2}) with $L_2$ loss (black dash) and  $L_1$ loss (blue dash-dot).  
 {\textit{Bottom:}} linearly interpolated noiseless outputs (solid line), and noisy measurements $(\circ)$ under nominal conditions (left) and in presence of outliers (right). The $L_1$ estimator performs much better in the presence of outliers.} 
    \label{FigIE}
     \end{center}
\end{figure*}
Classic approaches build parametric models 
of different orders using autoregressive (moving average) models  
with exogenous inputs AR(MA)X, 
fit to data using
Prediction Error Methods (PEM) 
\cite{Ljung,Soderstrom}. The ``best" model is selected using
complexity measures such as Akaike information criterion (AIC), Bayesian information criterion (BIC) 
or by cross validation (CV) techniques \cite{Akaike1974,schwarz1978estimating,Hastie01}.\\
This approach has a number of limitations~\cite{SS2010,PitfallsCV12}.  
For example, when the number of available output measurements 
is relatively small (so that asymptotic theory underlying AIC is not applicable), the
selected models often have poor predictive capability on new data.
In some cases, identifying a system is a highly ill-conditioned inverse problem
\cite{Bertero1}, and principled regularization techniques are required for estimation. 
The classical approach cannot incorporate additional information 
about the system, including domain restrictions, monotonicity, and unimodality;
this information can dramatically improve estimation. 
Finally, reliance on least-squares leaves the system identification process vulnerable 
to model mis-specification, including outliers in the observations 
~\cite{Hub,Gao2008,Aravkin2011tac,Farahmand2011}. 
 \\
{\bf Illustrative example}. Some of the drawbacks to the standard approach 
are illustrated by the example 
in Fig.~\ref{FigIE} (implementation details are given in Section~\ref{Sec4}). 
The impulse response (top panels, solid red line) is estimated from noisy outputs obtained using
white noise as system input. Consider two different situations. In the first case, 
the data set of 1000 samples is corrupted by white stationary Gaussian noise
(bottom left panel). The estimate obtained using the stable spline estimator 
(discussed in Section~\ref{SS+L2})
using the $L_2$ loss (top left, black dashdot line)  is close to the true signal. 
In the second case, the data set is corrupted by a few outliers 
(bottom right panel). Now, the impulse response profile 
(top right panel, black dashdot line)
reveals the vulnerability of the $L_2$ loss to deviations in the noise model.\\
\blue{
{\bf Contributions}. We build a modeling framework for LTI system identification, together with an open source solver (IPsolve)\footnote{https://github.com/saravkin/IPsolve} to fit all of the models of interest.  
The modeler can choose from a range of convex {\it piecewise linear-quadratic (PLQ)} 
penalties  
to use as misfit and regularizer. 
A particularly effective regularization strategy uses the {\it stable spline kernel} approach to control model space complexity. 
The modeler can also incorporate constraints on the signal response, significantly narrowing 
the search when additional information is available.  We now briefly describe each component of the  
proposed framework. }\\
{\bf Piecewise linear-quadratic penalties (PLQ)}. The limitations of $L_2$ motivate 
outlier-robust losses, including 
the $L_1$-norm, Huber~\cite{Hub}, Vapnik~\cite{Vapnik98,Pontil98} and the
hinge loss \cite{Evgeniou99,Scholkopf00}.
All of these penalties fall into the PLQ class, and appear in Figures~\ref{fig:quadratic}-\ref{fig:enet}.
Just as the $L_2$ penalty corresponds to maximum likelihood estimation with Gaussian errors, 
other penalties can be viewed as negative log likelihoods for non Gaussian noise \cite{NoiseG1991}. 
The $L_1$ loss corresponds to assuming the noise follows a Laplacian distribution~\cite{Aravkin2011tac}, 
and analogous likelihood interpretations have been developed for 
Vapnik and Huber penalties~\cite{AravkinBurkePillonetto2013}. 
\\
{\bf Stable-spline kernels}. Recent approaches cast system identification as a kernel learning problem,
formulated in a Hilbert space \cite{SS2010,SS2011,SurveyKBsysid}. 
Ill-posedness and ill-conditioning are studied within a Gaussian regression framework 
\cite{Rasmussen}. The unknown impulse response is modeled
as a Gaussian process whose covariance
encodes available prior knowledge, and estimators proposed in \cite{SS2010,PillACC2010}
model covariances using {\it{stable spline kernels}}, 
which include information on regularity and exponential stability of the impulse response.
Stable spline estimators have significant advantages over the classical
approach, especially in terms of the quality of the model complexity selection~\cite{SurveyKBsysid}.\\
{\bf General regularizers}. Though quadratic penalization of stable spline coefficients 
has proved to be effective, other choices can yield dramatic improvements. For example, 
if the impulse response is expected to have many zero entries,
the inclusion of a sparsity promoting prior can significantly improve the quality of the estimator,
e.g. the Laplace prior, or $L_1$ loss, used in the LASSO \cite{Lasso1996}.
This leads to a weighted combination of norms 
in the spirit of the elastic net procedure \cite{EN_2005}.
Our framework includes these priors, allowing any 
PLQ penalty to be used as a regularizer or data misfit, 
see Figures~\ref{fig:quadratic}-\ref{fig:enet}.
\\
{\bf Incorporating constraints.} 
\blue{
Additional information can be incorporated using 
inequality constraints on the impulse response coefficients. 
{\it Nonnegativity} is often a consequence of physical considerations.
Relaxed systems  with {\it monotonic responses} are frequently encountered 
in reciprocal electrical networks and mechanical systems with negligible inertial phenomena~\cite{Willems72} 
see e.g. the impulse response Fig. \ref{FigIE}.
A third example is bolus-tracking magnetic resonance imaging (MRI)~\cite{MRI2},
where quantification of cerebral hemodynamics requires estimation of impulse responses
known to be {\it positive and unimodal}.  
In all of these examples, there is a wealth of prior 
knowledge about the signal, and incorporating this knowledge into system identification
can yield significant improvements (see Figures~\ref{FigIE2} and \ref{FigMRI1})}.\\
\blue{
{\bf IPsolve}. Convex optimization has become a standard tool in many applications, and general convex solvers such as the MATLAB package CVX \cite{CVX2} and TFOCS~\cite{TFOCS}
are important tools for prototyping and testing new ideas. 
However, these general tools are not competitive with solvers that exploit problem-specific structure. 
IPsolve strikes a good balance between generality and problem structure, and outperforms TFOCS in the system identification context. IPsolve can solve any PLQ problem, but is less general than TFOCS, which can in principle solve any convex problem. 
On the other hand, IPsolve is much more general than solvers targeted to specific problem classes 
(e.g. $L_1$ regularization of smooth problems, or convex quadratic solvers). 
Nevertheless, the numerical experiments show that it is still competitive with targeted solvers for system identification. In particular,  for system identification problems that can be formulated as support vector regression (SVR), we compare IPsolve to libSVM~\cite{chang2011libsvm},  a state of the art solver for this SVR.
Similarly, for sparse system identification, IPsolve is competitive with fast iterative soft thresholding (FISTA)~\cite{Beck2009}, an optimal first-order method designed for smooth problems with simple regularizers. 
In general, first-order methods are faster than interior point methods for large-scale composite problems~\cite{JMLR:v15:aravkin14a}.
However, both system identification problems and stable spline kernels can be ill-conditioned, 
and interior-point methods are well-suited to ill-conditioned problems, 
which explains the result. }
We extend prior general work in PLQ modeling and optimization~\cite{AravkinIFAC12,AravkinBurkePillonetto2013} 
by incorporating inequality constraints,
and develop convergence guarantees for the entire framework  
in Theorem~\ref{basic ipa theorem}. 
We then apply the constrained PLQ framework 
to the linear system identification scenario,
and compare resulting estimators 
with classical PEM and stable spline approaches in 
a range of numerical studies, featuring contaminated data, and the inclusion of
additional information about the impulse response, e.g. unimodality or complete monotonicity. \\
{\bf Road map}. In Section \ref{Sec2}, we review the classical
approach to linear system identification, and provide a brief introduction to 
the stable spline estimation technique. 
In Section \ref{Sec3}, we formulate the general class of nonsmooth stable spline estimators,
using PLQ penalties as misfits and regularizers, and incorporate inequality constraints
on the impulse response. 
\blue{
We develop a general IP method for the class 
in Section \ref{IP section}, along with specific results for the system identification setting. 
In Section \ref{Sec4}, we compare against TFOCS, libSVM and FISTA to illustrate scaling and efficiency of IPsolve, and also test the performance of new estimators using several
Monte Carlo studies, including estimators for the example discussed in the introduction. 
While most of the experiments focus on the regime $n<<m$, 
in subsection \ref{SparseStable} we develop a sparse and stable estimator for the high-dimensional setting
($n>>m$) that arises in identifying multiple input single output (MISO) systems. 
This approach can be used to simultaneously identify connectivity and estimate models
in dynamic networks\cite{ChiusoSparse}.}

 

\section{PEM and stable spline approaches to linear system identification}
\label{Sec2}
%
%
%
%
Consider the following linear time-invariant discrete-time system 
\begin{equation} \label{MeasMod}
z(t) = G(q)u(t)+e(t)
\end{equation}
where $z\in \mathbb{R}^m$ is the output , $q$ is the shift operator 
\(
qu(t)=u(t+1),
\) 
$G(q)$ is the linear operator associated with the true system, assumed stable, 
$u$ is the input, and $e$ is white noise of variance $\sigma^2$.
Our problem is to estimate the system
impulse response assuming that the system input is known for 
$m$ measurements of $z$ at instants $t=1,\ldots,m$.
We then measure the quality of an estimator
$\widehat{G}$ by means of the fit measure
\begin{equation}\label{eq:fit}
\F(G,\hat G) = 100 \left(1- \| G -\widehat{G} \|_2/\| G \|_2   \right)
\end{equation} 
where, given a linear system $S(q)$, $\| S \|_2$
is the $L_2$-norm of its impulse response.
%
%

\noindent
The classic approach to system identification represents a parametrized model space $\mathcal{M}$ for linear systems
by a transfer function $G$ from input to output, parameterized by $x$:
\begin{equation}
  \label{eq:GH}
  z(t)=G(q,x)u(t)+e(t).
\end{equation} 
For instance, a standard \emph{black box} 
description assumes $G$ is a rational function of the shift
operator $q$,
\begin{equation}
  \label{eq:BB}
  G(q,x)=B(q)/C(q)
\end{equation} 
where $B(q)$ and $C(q)$ are polynomials in $q^{-1}$
whose unknown coefficients are the components of $x \in\mathbb{R}^n$. 
Different model structures can   
be associated with different degrees of $B(q)$ and $C(q)$.
For each model structure, the state $x$ 
can be estimated by PEM \cite{Ljung}, i.e. 
\begin{equation}
  \label{eq:5}
  \hat x = \argmin_x V (x), \quad   V(x)=\sum^m_{t=1}\big(z(t)-G(q,x)u(t) \big)^{2}.
\end{equation}
where the quadratic loss $V$ is a standard choice. 
In real applications, a suitable model structure (dimension of
$x$) is typically unknown and needs to be inferred from data.
This step is crucial, as it balances
bias and variance, and 
popular approaches include cross validation~\cite{Hastie01}, 
Akaike's criterion~\cite{Akaike1974}, and its small-sample version, 
corrected Akaike's criterion (AICc)~\cite{Hurvich}.


\subsection{The stable spline estimator}\label{SS+L2}
A drawback to rational transfer functions is that
they require solving (\ref{eq:5}), a
nonconvex and potentially high-dimensional problem, for each postulated model order. 
A common alternative is the finite impulse response (FIR) model obtained setting $C(q)=1$
in (\ref{eq:BB}), which makes (\ref{eq:5}) a linear least-squares problem in 
the polynomial coefficients $x$ for $B$:
\begin{equation}
\label{eq:transition}
\begin{aligned}
V(x) & = \sum^m_{t=1}\big(z(t)-B_x(q)u(t) \big)^{2}  = \sum_{t=1}^m (z_t - \langle \phi(u_t,q), x\rangle)^2,
\end{aligned}
\end{equation}
where $\phi(u_t,q)$ is a vector determined by the input and shift operators. 
However, a high-order FIR, often necessary to capture system dynamics,
can suffer from high variance, so regularization is crucial.\\
We quickly review the regularized approaches described in \cite{SS2010,ChenOL12}.  
 First rewrite the measurement model
(\ref{eq:transition}) using matrix-vector notation:
\begin{equation}\label{MatrixModMV}
  z=\Phi x + E,   
\end{equation}
where $z \in \mathbb{R}^m$ is a vector comprising the $m$ output measurements, 
$E$ is the noise,
$x \in \mathbb{R}^n$ is the (column) vector 
of impulse response coefficients,
and $\Phi$ is a matrix with rows $\phi_t = \phi(u_t,q)$ determined by input values and shift operator.
 {For instance, assuming an input delay of one sample, we have 
 $$
 z=\left(\begin{array}{c}z_1 \\ z_2 \\ \vdots \\ z_m \end{array}\right), \ \ 
 \Phi=\left(\begin{array}{cccc} u_0 & u_{-1} & \ldots & u_{-n+1} \\
 u_1 & u_{0} & \ldots & u_{-n+2}  \\
 \vdots & \vdots & \vdots & \vdots \\
 u_{m-1} &  u_{m-2} & \ldots &  u_{-n+m} \end{array}\right), \ \  x=\left(\begin{array}{c}x_1 \\ x_2 \\ \vdots \\ x_n \end{array}\right).
 $$
 \color{red} }
In contrast to classical
approaches to system identification, the $n$th-order FIR 
approach does not need to balance  
bias and variance, but only needs to be of sufficiently large order to capture the system 
dynamics. The model complexity is controlled via stable spline kernels.
A stable spline estimator for impulse response solves
\begin{equation}
\label{eq:MV}
  \hat x = \arg \min_{x} \|z-\Phi x\|^2  + \gamma x^T Q^{-1} x\;,
\end{equation}
where 
the positive scalar $\gamma$ is the {\it regularization parameter},
while $Q \in \mathbb{R}^{n \times n}$ is a regularization matrix defined by the class of the stable
spline kernels \cite{PillACC2010}. 
{The choice of $\gamma$ is important:  
ideally, it must be tuned so that a small bias is introduced in the estimation process in order to significantly reduce the variance
(in comparison with a baseline such as least squares).}

Problem (\ref{eq:MV}) is always well-posed because of the strongly convex term $\gamma x^T Q^{-1} x$,
which also controls model order. 
When using the discrete-time version 
of the first-order stable spline kernel (also called TC kernel in  \cite{ChenOL12}),   
the $(i,j)$ entry of $Q$ is specified to be
\begin{equation}\label{eq:TC}
Q_{ij} := \alpha^{\max(i,j)}, \quad \mbox{for some $0 \leq \alpha <1$}\;.
\end{equation} 
Smoother impulse response estimates can be obtained by  
using the second-order stable spline kernel. In this case, 
the entries of $Q$ are given by
\begin{equation}\label{eq:SS}
Q_{ij} := \left[ \alpha^{ (i+j)}
\alpha^{\max(i,j)}/2-\alpha^{3 \max(i,j)}/6\right], \quad \mbox{for some $0 \leq \alpha <1$}\;.
\end{equation} 

\blue{The kernel Q may be ill-conditioned (the condition number  
grows to infinity with small $\alpha$ and large $n$). Stability properties of first-order stable-spline 
kernels, as well as formulas for their inverses and Cholesky factors are discussed in detail in~\cite{carli2014maximum}.
While ill-conditioning of the kernel can partially controlled by hyper-parameter selection,
general system identification problems may be ill-conditioned, and the implications 
are discussed in Section~\ref{Sec4}. 
} 

\blue{
In (\ref{eq:TC}) and (\ref{eq:SS}), $\alpha$ is a kernel hyperparameter related
to the dominant pole of the system (i.e. it establishes how fast the impulse response
decays to zero) and is typically unknown.
%
The estimator (\ref{eq:MV}),where $Q$
is given by (\ref{eq:TC}) or (\ref{eq:SS}), depends on 
$\alpha$ and $\gamma$, which need to be determined from data
using marginal likelihood maximization \cite{Maritz:1989,MacKay,BergerBook,AravkinIFAC12,JMLR:v15:aravkin14a}.
%
Learning hyperparameters is analogous to model order selection in the classical
PEM framework. Once the two parameters are found, the impulse response estimate 
for the least-squares formulation can be obtained by solving a linear system of equations
given by the first-order optimality conditions for the problem (\ref{eq:MV}).} \\

\section{New formulations of the stable spline estimator}\label{Sec3}
\label{sec:PLQs}

We now introduce the class of piecewise linear quadratic (PLQ)
functions and penalties. We first develop a representation calculus 
for estimators of interest using simple PLQ building blocks, 
and then show how to formulate general estimation problems 
as minimizers of a single PLQ objective over a polyhedral set. 
All estimators are obtained 
using the algorithm developed in Section \ref{IP section}.

\subsection{From quadratic to PLQ penalties}
\label{simpleExamples}
It is useful
to rewrite the estimator (\ref{eq:MV}) as:
\begin{equation}
\label{SysIdEqs}
y = L^{-1}x, \quad Q = LL^T
\end{equation}
where $L$ is invertible thanks to the 
positive definite property of stable spline kernels 
(we caution that the variable $y$ introduced here is \emph{not} the same object
as the function $y(t)$ defined in (\ref{MeasMod})). 
In the new variable $y$ the estimation problem (\ref{eq:MV}) translates into the problem
\begin{equation}
\label{SysIdObjectiveTwo} 
\min_y 
\left\|z - \Phi Ly\right\|^2+ \gamma \|y\|^2\;.
\end{equation}
This estimator uses quadratic functions
for both the misfit penalty and the regularizer.  
The goal of the remainder of the paper is to show how to generalizing
\eqref{SysIdObjectiveTwo} for adaptation to a variety of data scenarios and to
illustrate the computational efficacy of these adaptations.

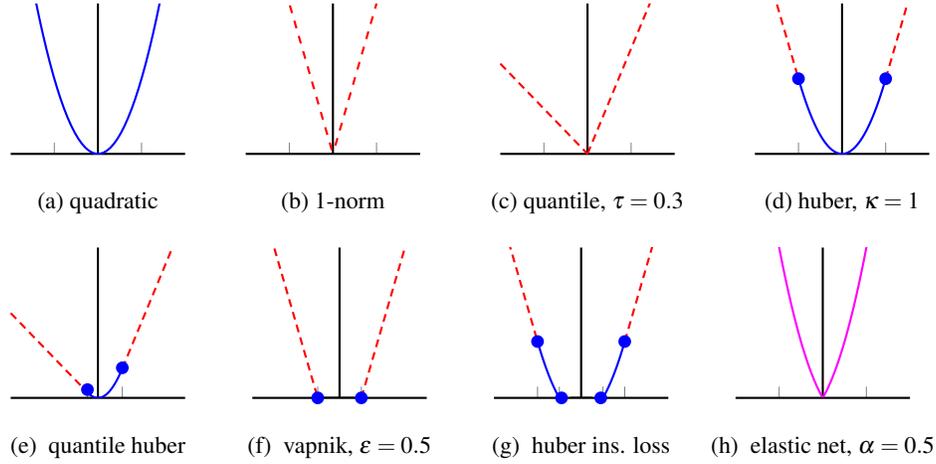
\begin{figure*}[t!]
    \begin{subfigure}[t]{0.24\textwidth}
       \centering
\begin{tikzpicture}
  \begin{axis}[
    thick,
    height=2cm,
    xmin=-2,xmax=2,ymin=0,ymax=1,
    no markers,
    samples=50,
    axis lines*=left, 
    axis lines*=middle, 
    scale only axis,
    xtick={-1,1},
    xticklabels={},
    ytick={0},
    ] 
\addplot[blue, domain=-2:+2]{.5*x^2};
  \end{axis}
  \end{tikzpicture}
  \caption{\label{fig:quadratic}quadratic}
    \end{subfigure}%
    \begin{subfigure}[t]{0.24\textwidth}
        \centering
\begin{tikzpicture}
  \begin{axis}[
    thick,
    height=2cm,
    xmin=-2,xmax=2,ymin=0,ymax=1,
    no markers,
    samples=100,
    axis lines*=left, 
    axis lines*=middle, 
    scale only axis,
    xtick={-1,1},
    xticklabels={},
    ytick={0},
    ] 
  \addplot[red, dashed, domain=-2:+2]{abs(x)};
  \end{axis}
\end{tikzpicture}
    \caption{\label{fig:1norm}1-norm}
    \end{subfigure}
    \begin{subfigure}[t]{0.24\textwidth}
        \centering
\begin{tikzpicture}
  \begin{axis}[
    thick,
    height=2cm,
    xmin=-2,xmax=2,ymin=0,ymax=1,
    no markers,
    samples=50,
    axis lines*=left, 
    axis lines*=middle, 
    scale only axis,
    xtick={-1,1},
   xticklabels={},
    ytick={0},
    ] 
\addplot[red,domain=-2:0,densely dashed]{-.3*x};
\addplot[red,domain=0:+2,densely dashed]{.7*x};
  \end{axis}
\end{tikzpicture}
            \caption{\label{fig:quantile}quantile, $\tau = 0.3$}
        \end{subfigure}
   \begin{subfigure}[t]{0.24\textwidth}
   \centering
   \begin{tikzpicture}
  \begin{axis}[
    thick,
    height=2cm,
    xmin=-2,xmax=2,ymin=0,ymax=1,
    no markers,
    samples=50,
    axis lines*=left, 
    axis lines*=middle, 
    scale only axis,
    xtick={-1,1},
    xticklabels={},
    ytick={0},
    ] 
\addplot[red,domain=-2:-1,densely dashed]{-x-.5};
\addplot[blue, domain=-1:+1]{.5*x^2};
\addplot[red,domain=+1:+2,densely dashed]{x-.5};
\addplot[blue,mark=*,only marks] coordinates {(-1,.5) (1,.5)};
  \end{axis}
\end{tikzpicture}
\caption{\label{fig:huber}huber, $\kappa = 1$}
\end{subfigure}    
   \begin{subfigure}[t]{0.24\textwidth}
   \centering
\begin{tikzpicture}
  \begin{axis}[
    thick,
    height=2cm,
    xmin=-2,xmax=2,ymin=0,ymax=1,
    no markers,
    samples=100,
    axis lines*=left, 
    axis lines*=middle, 
    scale only axis,
    xtick={-.24,.56},
    xticklabels={},
    ytick={0},
    ] 
\addplot[red,domain=-2:-2*0.3*0.4,densely dashed]{0.3*abs(x) - 0.4*0.3^2};
\addplot[blue,domain=-2*0.3*0.4:2*(1-0.3)*0.4]{0.25*x^2/0.4};
\addplot[red,domain=2*(1-0.3)*0.4:2,densely dashed]{(1-0.3)*abs(x) - 0.4*(1-0.3)^2};
\addplot[blue,mark=*,only marks] coordinates {(-.24,0.0550) (0.56,0.20)};
  \end{axis}
\end{tikzpicture}
\caption{\label{fig:huberQ} quantile huber}
\end{subfigure}  
   \begin{subfigure}[t]{0.24\textwidth}
   \centering
\begin{tikzpicture}
  \begin{axis}[
    thick,
    height=2cm,
    xmin=-2,xmax=2,ymin=0,ymax=1,
    no markers,
    samples=50,
    axis lines*=left, 
    axis lines*=middle, 
    scale only axis,
    xtick={-0.5,0.5},
    xticklabels={},
    ytick={0},
    ] 
    \addplot[red,domain=-2:-0.5,densely dashed] {-x-0.5};
    \addplot[domain=-0.5:+0.5] {0};
    \addplot[red,domain=+0.5:+2,densely dashed] {x-0.5};
    \addplot[blue,mark=*,only marks] coordinates {(-0.5,0) (0.5,0)};
  \end{axis}
\end{tikzpicture}
\caption{\label{fig:vapnik} vapnik, $\epsilon = 0.5$}
\end{subfigure}  
   \begin{subfigure}[t]{0.24\textwidth}
   \centering
\begin{tikzpicture}
  \begin{axis}[
    thick,
    height=2cm,
    xmin=-2,xmax=2,ymin=0,ymax=1,
    no markers,
    samples=50,
    axis lines*=left, 
    axis lines*=middle, 
    scale only axis,
    xtick={-1,1, -.5, .5},
    xticklabels={},
    ytick={0},
    ] 
\addplot[domain=-0.25:+0.25] {0};
\addplot[red,domain=-2:-1,densely dashed]{-x-.5-.5*.25};
\addplot[blue, domain=-1:-.25]{.5*x^2-.5*.25};
\addplot[blue, domain=.25:1]{.5*x^2-.5*.25};
\addplot[red,domain=+1:+2,densely dashed]{x-.5-.5*.25};
\addplot[blue,mark=*,only marks] coordinates {(-1,.5-.5*.25) (1,.5-.5*.25)(-.45, 0) (.45, 0)};
  \end{axis}
\end{tikzpicture}
\caption{\label{fig:sel} huber ins. loss}
\end{subfigure}  
   \begin{subfigure}[t]{0.24\textwidth}
   \centering
\begin{tikzpicture}
  \begin{axis}[
    thick,
    height=2cm,
    xmin=-2,xmax=2,ymin=0,ymax=1,
    no markers,
    samples=100,
    axis lines*=left, 
    axis lines*=middle, 
    scale only axis,
    xtick={-1,1},
    xticklabels={},
    ytick={0},
    ] 
\addplot[amethyst, domain=-2:+2]{.5*x^2 + 0.5*abs(x)};
  \end{axis}
\end{tikzpicture}
\caption{\label{fig:enet} elastic net, $\alpha = 0.5$}
\end{subfigure}  
    \caption{Common piecewise linear-quadratic (PLQ) losses.}
\end{figure*}
Let $Y$ denote the feasible polyhedral constraint region for $y$. 
Then an explicit representation for $Y$ can be written as
\begin{equation}
\label{Yrepresentation}
Y = \{y: A^Ty \leq a\}\;, \quad A\in \mathbb{R}^{n \times p}, \quad a\in\mathbb{R}^{p}.
\end{equation}
This allows us to represent prior knowledge about the signal, including 
domain information (e.g. lower and upper limits), as well as monotonicity or unimodality properties.\\ 
We consider generalizations of~\eqref{SysIdObjectiveTwo} that use any PLQ penalty: 
\begin{equation} 
\label{probTwo}
\min_{y \in Y}   V \left(z - \Phi Ly \right)  + \gamma W\left(y\right),
\end{equation}
where $V$ and $W$
are piecewise linear quadratic functions
introduced below, and $Y$ is as 
in~\eqref{Yrepresentation}. 
Nine important examples of 
these penalties appear in Figures~\ref{fig:quadratic}-\ref{fig:enet}.
\begin{table}
\blue{
\caption{\label{table:PLQ} Dual representations of common PLQ penalties.}
\begin{tabular}{|c|c|c|}\hline
Penalty & Representation~\eqref{PLQpenalty} & Selected references\\ \hline
Quadratic, Fig.~\ref{fig:quadratic} 
& $ \sup_{u} \left\{ux - \frac{1}{2}u^2\right\}$ 
& \cite{freedman2009statistical,SeberWild2003}\\
\hline
1-norm, Fig.~\ref{fig:1norm} 
&$ \sup_{u \in [-1,1]} \left\{ux\right\}$ 
& \cite{Hastie90,LARS2004,Donoho2006,Elad:08}\\
\hline
Quantile, Fig.~\ref{fig:quantile} 
&$ \sup_{u \in [-\tau,(1-\tau)]} \left\{ux\right\}$ 
& \cite{KB78,KG01,Koenker:2005}\\
\hline
Huber, Fig.~\ref{fig:huber} 
&$\sup_{u\in[-\kappa, \kappa]}\left\{ ux - u^2/2\right\}$ 
& \cite{Hub,Mar,LiW98}\\
\hline
Q-Huber, Fig.~\ref{fig:huberQ} 
&$ \sup_{u \in [-\kappa\tau,\kappa(1-\tau)]} \left\{ux - u^2/2\right\}$ 
& \cite{aravkin2014qh}\\
\hline
Vapnik, Fig.~\ref{fig:vapnik} 
&$ \displaystyle\sup_{u \in [0,1]^2}
\left\langle 
\begin{bmatrix}1  \\ 
-1 \end{bmatrix}x 
-
\begin{bmatrix}\epsilon \\ 
\epsilon \end{bmatrix} , 
u \right\rangle $ 
& \cite{Vapnik98,Hastie01,Scholkopf00,AravkinIFAC}\\
\hline
SEL, Fig.~\ref{fig:sel} 
&$ 
\displaystyle
\sup_{u \in [0,1]^2}
\left\{
\left\langle 
\begin{bmatrix}1  \\ 
-1 \end{bmatrix}x 
-
\begin{bmatrix}\epsilon \\ 
\epsilon \end{bmatrix} , 
u \right\rangle - u^2/2
\right\}$ 
& \cite{chu2001unified,lee2005epsi,dekel2005smooth}\\
\hline
Elastic net, Fig.~\ref{fig:enet} 
&$ 
\displaystyle
\sup_{u \in [0,1] \times \mathbb{R}}
\left\{
\left\langle 
\begin{bmatrix}1  \\ 
1 \end{bmatrix}x , u \right\rangle
 - u_2^2/2\right\}$ 
& \cite{ZouHuiHastie:2005,EN_2005,li2010bayesian,de2009elastic}\\
\hline
\end{tabular}
}
\end{table}
\blue{
\begin{definition}[PLQ functions and penalties]
\label{generalPLQ}
A piecewise linear quadratic (PLQ) function is any function 
$\rho(c, C, b, B, M; \cdot): \mB{R}^n \rightarrow \mathbb{\overline R}$ 
admitting representation
\begin{equation}\label{PLQpenalty}
\rho(c, C, b, B, M; y) 
=
\sup_{u \in U} 
\left\{      \langle u,b + By \rangle -  \frac{1}{2} \langle u, Mu \rangle
 \right\} \;,
\end{equation}
where $U = \{u: C^Tu \leq c\}$ is a polyhedral set containing the origin, 
$M$ is symmetric positive semidefinite, $\in \mathbb{R}^k$,
$B\in\mB{R}^{k\times n}$ with $null(B) = \{0\}$. 
\end{definition}\\
}
%
%
The eight loss functions illustrated in Figures~\ref{fig:quadratic}-\ref{fig:enet} are members of
the PLQ class; dual representations~\eqref{PLQpenalty} and references are given in Table~\ref{table:PLQ}.
In the following section, we use these representations to develop fast algorithms. 

\blue{
{\bf Modeling with PLQ.} We can classify PLQs according to three features: behavior at origin, symmetry, and tail growth. 
Several situations are considered in the simulation studies. \\
{\bf Origin.} Nonsmooth behavior at origin promotes sparsity. When used as a regularizer, the 1-norm and quantile loss find sparse solutions. 
When used as a misfit, these losses fit some of the data exactly, quadratic behavior at the origin 
will fit data approximately, and the Vapnik misfit corresponds to uniform residuals.\\
{\bf Tail growth}. Tail growth allows robustness to outliers. Vapnik, $1$-norm, and Huber all have similar robustness properties. 
The quadratic and elastic net losses are not robust to outliers. \\
{\bf Symmetry}. Asymmetric losses model cases where either (a) positive or negative responses 
are more likely (asymmetric regularizer), or (b) costs for over-estimating or under-estimating observations are different (asymmetric loss). 
}
%

The representation in Definition~\ref{generalPLQ} is explicitly used
to solve the generalized linear system identification problem (\ref{probTwo}).
We first derive a PLQ representation calculus. 

\begin{remark}[Affine composition]
\label{affineComposition}
Take any PLQ function $\rho(c, C, b, B, M; y)$. Suppose that $y = Ex + e$,
where $x\mapsto Ex+e$ is an injective affine transformation in $x$. 
Then we have 
\[
\rho(c, C, b, B, M; Ex + e) = \rho(c, C, b + Be, BE, M; x)
\] 
so the composition is also a PLQ function, with representation 
$c, C, b + Be, BE, M$. 
\end{remark}
\begin{remark}[PLQ addition]
\label{plqAddition}
Given two PLQ functions $\rho(c_1, C_1, b_1, B_1, M_1; y)$ and $\rho(c_2, C_2, b_2, B_2, M_2; y)$, 
the sum is also a PLQ function, with representation 
\[
\begin{aligned}
c \!\!=\!\!\! \begin{bmatrix} c_1  \\ c_2\end{bmatrix},\; 
C \!\!=\!\!\! \begin{bmatrix}C_1 & 0 \\ 0 & C_2 \end{bmatrix},\;  
b \!\!=\!\!\! \begin{bmatrix} b_1  \\ b_2\end{bmatrix},\; 
B \!\!=\!\!\! \begin{bmatrix} B_1  \\ B_2\end{bmatrix},\; 
M \!\!=\!\!\! \begin{bmatrix}M_1 & 0 \\ 0 & M_2 \end{bmatrix}.
\end{aligned}
\]
\end{remark}
The PLQ class is closed under addition and affine composition,
allowing the design of a PLQ penalty that is well suited to a given application.
For given PLQ penalties $V$ and $W$, their sum~\eqref{probTwo} is also a PLQ
penalty, with a representation that can be automatically constructed from individual components using the above 
remarks. Once a representation for~\eqref{probTwo} is constructed, can optimize it over any polyhedral set, 
as shown in the next section.

\section{An Interior Point (IP) Approach}\label{IP section}

We now show how to solve 
\begin{equation}
\label{genELQP}
\begin{aligned}
\min_{y}\quad \rho(c, C, b, B, M; y)  \quad \text{s.t.} \quad A^Ty \leq a,
\end{aligned}
\end{equation}
using interior point IP methods~\cite{KMNY91,NN94,Wright:1997}. 
IP methods solve nonsmooth optimization problems 
by applying a damped Newton method to a homotopy path that
parametrizes the underlying Karush-Kuhn-Tucker (KKT) system.
In this regard,
the first key observation is that the KKT system for \eqref{genELQP} 
is an instance of a 
{\em monotone mixed linear complementarity problem} (MLCP) \cite{Wright:1997}
since it can be written as
\begin{equation}\label{MLCP}
\begin{pmatrix} s\\ r\\ 0\\ 0\end{pmatrix}=
\left[
\begin{array}{cc|cc}
0&0&-C^T&0\\
0&0&0&-A^T\\
\hline
C&0&M&-B\\
0&A&B^T&0
\end{array}
\right]
\begin{pmatrix} q\\ w\\ u\\ y\end{pmatrix} +
\begin{pmatrix} c\\ a\\ b\\ 0\end{pmatrix}
\end{equation}
with 
\begin{equation}\label{MLCP complementarity}
0\le \begin{pmatrix}q\\ w\end{pmatrix},\ \begin{pmatrix}s\\ r\end{pmatrix}\mbox{ and }
\begin{pmatrix}q\\ w\end{pmatrix}^T\begin{pmatrix}s\\ r\end{pmatrix}=0\ ,
\end{equation}
where the matrix in \eqref{MLCP} is positive semi-definite
(see Appendix for details).
Consequently, it is possible to transform this MLCP into a monotone LCP and solve it
by an interior point algorithm \cite{KMNY91}. 
However, this transformation is arduous,
especially in high dimensions, and may be prohibitively expensive \cite{ALP97,Gul95,Wri96}. 
In \cite{Wri96} it is noted that the transformation to an LCP is not essential
if the matrix
\begin{equation}\label{MLCP2}
\mathcal{H}:=
\begin{bmatrix}
-C^T&0\\
0&-A^T\\
M&-B\\
B^T&0
\end{bmatrix}\ \in\ \mathbb{R}^{(\ell+p+k+n)\times (k+n)}
\end{equation}
is injective.
In our context, the injectivity of this matrix can be established under mild conditions.

\begin{theorem}[Injectivity of $\mathcal{H}$]\label{injectivity}
Suppose $M$ is symmetric positive semidefinite and $null(B) = \{0\}$.
Then the matrix $\mathcal{H}$ in \eqref{MLCP2} is injective if and only if 
\begin{equation}\label{injectivity condition}
\Nul{M}\cap\Nul{B^T}\cap\Nul{C^T}=\{ 0\}\, .
\end{equation}
\end{theorem}
%
Condition \eqref{injectivity condition} is satisfied if the stronger condition
\begin{equation}\label{simple ic}
\Nul{M}\cap\Nul{B^T}=\{0\}
\end{equation}
holds. 
This latter condition is satisfied by all of the PLQ functions in 
Section \ref{simpleExamples}~\cite{AravkinBurkePillonetto2013,AravkinBurkePillonetto2013b}.\\
%
We first  specify the dimensions of the quantities appearing in
\eqref{genELQP}. Let 
\begin{equation}\label{details nailed down}
\begin{aligned}
b\in \mathbb{R}^k,\, C \in \mathbb{R}^{k \times l}, \, 
c\in \mathbb{R}^l,\,  
B \in \mathbb{R}^{k\times n}, \; 
A \in \mathbb{R}^{n\times p}, \, M \in \mathbb{R}^{k\times k},\, 
\mbox{ and }a \in \mathbb{R}^p,
\end{aligned}
\end{equation} 
and set $N:=2l+2p+k+n$.
Then, 
given $\mu\ge0$, define 
$\map{F_\mu}{\mathbb{R}^N}{\mathbb{R}^N}$ by
%
\begin{equation}\label{F mu}
F_\mu(q,w,u,y,s, r) := 
\left(
\begin{matrix}
C^Tu + s - c\\
A^Ty + r - a\\
Mu + Cq -By - b \\
B^Tu  + Aw\\
Qs - \mu \B{1} \\
Wr - \mu\B{1}
\end{matrix}
\right) \;,
\end{equation}
where $Q:=\mbox{diag}(q)$ and $W:=\mbox{diag}(w)$.
The KKT conditions \eqref{MLCP}-\eqref{MLCP complementarity} 
are
\begin{equation}
\label{F mu KKT}
F_0(q, w,u,y,s, r)=0\mbox{ for } 
s,q\in\mathbb{R}^l_+\mbox{ and }
r,w\in\mathbb{R}^p_+.
\end{equation}
The variables
$y\in\mathbb{R}^n$ and $u\in\mathbb{R}^k$ 
are those that appear in the definition of the PLQ function $\rho$
\eqref{PLQpenalty}, 
$s$ and $r$ are  {\it slack variables},  
and $q$, $w$ are the dual variables 
that correspond to constraints $C^Tu\le c$ and $A^Ty\le a$.
For any positive integer $\ell$, we set
\(
\mB{R}^\ell_+:=\{x\in\mB{R}^\ell\, :\, 0\le x_i,\, i=1,2,\dots,\ell\}
\) and denote the interior of \(\mB{R}^\ell_+\) by \(\mB{R}^\ell_{++}\).

An interior point approach applies a damped Newton iteration to a relaxed
version of the KKT system by solving \eqref{F mu KKT} for $\mu>0$ and letting 
$\mu$ carefully descend to zero. 
We choose an initial
$(y^0,u^0)\in \mB{R}^n\times\mB{R}^k$ and
$(s^0,r^0,q^0,w^0)\in \mathcal{D}_{++}$
where 
$ \mathcal{D}_{++}:=\mB{R}^l_{++}\times\mB{R}^p_{++}\times\mB{R}^l_{++}\times\mB{R}^p_{++}$,
and then preserve positivity of the iterates $(s^\nu,r^\nu,q^\nu,w^\nu)$
at each iteration of the damped Newton method for~\eqref{F mu KKT}. 
For this to succeed, the Newton iteration must be well-defined; 
in particular \(F_\mu^{(1)}\) must be invertible at all iterates when $\mu>0$.
On 
$\mathcal{D}_{++}$, we have
\begin{equation}\label{F'}
F_\mu^{(1)}(q,w,u,y,s,r)=
\left[
\begin{array}{cc|cc|cc}
0&0&-C^T&0&-I&0\\
0&0&0&-A^T&0&-I\\
\hline
C&0&M&-B&0&0\\
0&A&B^T&0&0&0\\
\hline
Q&0&0&0&S&0\\
0&W&0&0&0&R
\end{array}
\right],
\end{equation}
where $S=\mathrm{diag}(s)$ and $R=\mathrm{diag}(r)$. 
We now show that the invertibility of \(F_\mu^{(1)}\)
is related 
to the condition \eqref{injectivity condition} in Theorem \ref{injectivity}.

\begin{theorem}[Invertibility of $F_\mu^{(1)}$]\label{invertibility}
Given $(u,y)\in\mB{R}^k\times\mB{R}^n$ and
$(q,w,s,r)\in  \mathcal{D}_{++}$, the matrix $F_\mu^{(1)}(q,w,u,y,s,r)$ is invertible
if and only if the matrix
\begin{equation}\label{reduced}
\begin{bmatrix}
M+CSQ^{-1}C^T&-B\\
B^T&ARW^{-1}A^T
\end{bmatrix}
\end{equation}
is invertible, which, in turn, is equivalent to condition \eqref{injectivity condition}.
\end{theorem}\\
Note that the central $2\times 2$ block of $F_\mu^{(1)}(q,w,u,y,s,r)$
is precisely the block matrix appearing in the lower right of the MLCP 
matrix \eqref{MLCP}, and Theorem~\ref{invertibility} relates~\eqref{injectivity condition} 
to the algebraic implementation of the interior point method for the general 
problem~\eqref{genELQP}, detailed in Algorithm~\ref{algor}. 
To simplify notation we let $\chi = \begin{bmatrix} q^T & w^T & u^T & y^T & s^T & r^T\end{bmatrix}^T$.

\begin{algorithm}[h!]
\caption{Interior Point Method for~\eqref{probTwo}.}
\label{algor}
\begin{algorithmic}[1]
\Require{$x^0, y^0, u^0, s^0, r^0, q^0, \eta \in (0,1), \mu^0\in \mathbb{R}_{++}, \mathrm{tol}>0$}
\While{not converged}
\Let{$d$}{$-(F_{\mu}^{(1)})^{-1}F_\mu$}
\Let{$t$}{$\max \; \tau \quad \mbox{s.t.}  \quad  \left\{ \begin{bmatrix} s + \tau d_s \\ r + \tau d_r\end{bmatrix},\; \begin{bmatrix} q + \tau d_q \\ w + \tau d_w \end{bmatrix}\right\} > 0, \; 
\|F_\mu( \chi + \tau d )\| \leq (1-\eta) \|F_\mu(\chi)$}
\Let{$\chi$}{$\chi + td$}
\Let{$\mu$}{$0.1(s^T q + r^T w)/(p+l)$}
\Let{converged}{$(\mu < \mathrm{tol})$}
\EndWhile 

\Return{$\chi$}
\end{algorithmic}
\end{algorithm}

Let $\tau> 0$ and define
\[
\mathcal{F}_+(\tau)\!\!:=\!\!\left\{\!\!(q,w,u,y,r,s) \!\!
\left|\!\!\begin{array}{c}
(u,y)\!\in\!\mB{R}^k\!\times\!\mB{R}^n,\, (q,w,s,r)\!\in\!  \mathcal{D}_{++}
\\
\mbox{ and equation \eqref{MLCP} is satisfied} 
\\
\mbox{with }q^Ts+w^Tr\le\tau
\end{array}\!\!\!
\right.\!\!\right\}
\]
and $\mathcal{C}:=\left\{(q,w,u,y,r,s) 
\left| 
\mbox{\eqref{F mu KKT} holds for some $\mu>0$}\right.\right\}\, $. 
The set $\mathcal{C}$ is called the {\em central path}. The key to the complexity
analysis for the algorithm is to ensure that the iterates hew sufficiently close to this 
path as $\mu$ descends to zero. 

\begin{theorem}[Convergence Properties]\label{basic ipa theorem}
Consider any optimization problem of the form~\eqref{genELQP}
satisfying \eqref{details nailed down}.
If $\mathcal{F}_+(+\infty)\ne\emptyset$ and \eqref{injectivity condition} holds,
then Algorithm \ref{algor} is implementable, 
the sets $\mathcal{F}_+(\tau)$ are
non-empty, convex, and bounded for all $\tau> 0$, the central path is well-defined,
and every cluster point of the central path as $\mu\downarrow 0$ is a KKT point
for \eqref{genELQP}.
\end{theorem}

The proof 
and details for computing the Newton step are given in the appendix.  
Matrices
\begin{equation}\label{T and Omega}
T:=M+CSQ^{-1}C^T\mbox{ and }\Omega:=B^TT^{-1}B+ARW^{-1}A^T
\end{equation}
and their inverses play a key role; 
$T$ is invertible since \eqref{simple ic} holds, and $\Omega$ is invertible
since $T$ is invertible and $B$ is injective.
The sparsity of these matrices determine the complexity of the algorithm, computing
 the Newton step is the main effort at each iteration. 
In all our PLQ examples, $M$ and $C$ are very sparse, and the matrix $T$ is  diagonal.
The next result describes the per iteration complexity of the algorithm in this setting. 



\begin{theorem}[PLQ Iteration Complexity]
\label{ipTheorem}
If the matrices $T_k:=M+CS_kQ_k^{-1}C^T$ in Algorithm \ref{algor}
are diagonal, 
then every interior point iteration can be computed with complexity 
\blue{
$O(\min(n,k+p)^2(k  + p  + n))$. 
}
\end{theorem}

\blue{
If $n < k+p$, we obtain the complexity above by forming and inverting $\Omega$ in~\eqref{T and Omega}. 
Otherwise, we apply the Sherman-Morrison-Woodbury formula to form and invert a matrix in dimension (k+p).
The choice is made by IPsolve based on the dimensions of the inputs. 
}
Turning our attention back to system identification, 
$n$ is the dimension of the impulse response, while $k$ and $l$ 
depend on $m$, in particular $k \geq m$, 
while $l$ depends on the structure of the PLQ penalties used to build 
the estimate~\eqref{probTwo}. 


\begin{corollary}[SysID Iteration Complexity]
\label{sysIDcor}
If the constraint matrix $A$ contains $O(n)$
entries (as e.g. with box constraints), 
while matrices $B$ and $C$ have on the order 
of $m$ entries, each interior point iteration can be solved 
with complexity \blue{$O(\min(m,n)^2(m+ n))$.}  
\end{corollary}

The above result shows that IP computational complexity 
scales favorably with the number of measurements $m$ which,  in 
system identification, is typically much larger than the number of 
unknown impulse response coefficients $n$.

\section{Numerical studies}
\label{Sec4}

The  new approach
is now tested using numerical studies.
\subsection{Introductory example: robust estimation with inequality constraints }
\label{IntEx}

\begin{figure}
  \begin{center}
\hspace{.1in}
 { \includegraphics[scale=0.25]{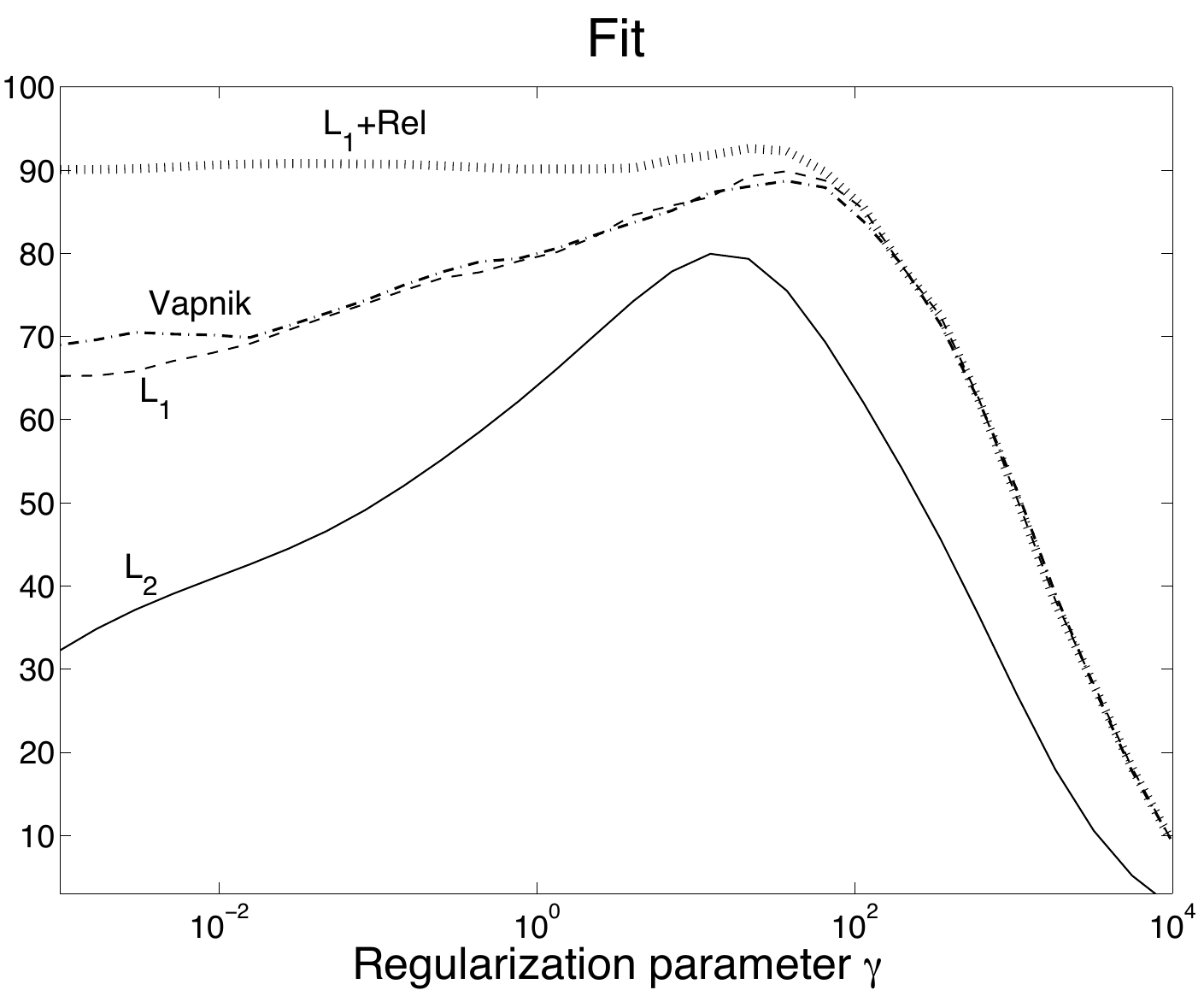}} 
 \caption{Introductory example. Fit, as a function of the regularization parameter $\gamma$,
 obtained using the stable estimator equipped with the $L_2,L_1$ and Vapnik loss,
 and combining the $L_1$ loss with the information that data come from a relaxation system ($L_1+Rel$).} 
    \label{FigIE2}
     \end{center}
\end{figure}

Results in Fig. \ref{FigIE} illustrate the well established fact that the estimator $SS+L_2$ 
based on the quadratic loss is
vulnerable to outliers
(recall that $SS$ refers to stable spline in Section~\ref{SS+L2}). 
Here we exploit the general framework of the previous section
to design a robust estimator by replacing the quadratic (Gaussian) with the absolute value 
(Laplace) loss.
This leads to the estimator $SS+L_1$ defined by
\begin{equation}\label{SS+L1}
\hat x = \arg \min_{x} \|z-\Phi x\|_1  + \gamma x^T Q^{-1} x\;.
\end{equation}
This objective can be transformed into form~\eqref{probTwo} 
and then into \eqref{genELQP} as described in Section \ref{sec:PLQs}.
As in the quadratic case (\ref{eq:MV}), the solution depends on the unknown parameters 
$\gamma$ and $\alpha$ (which enters $Q$). To solve \eqref{SS+L1},
$\gamma$ and $\alpha$ are estimated via cross validation, splitting the data 
into a training and a validation set of equal size. 
The ``optimal" hyperparameters values 
are obtained by searching over a two dimensional grid. In particular, 
$\alpha$ and $\gamma$ assume values on the grid defined,
respectively, by the MATLAB commands 
\begin{verbatim}
 			A=[0.01 0.05:0.05:0.95 0.99], 
 			B=logspace(log10(g/100),log10(g*100),50),
\end{verbatim}
with \texttt{g} set to the value of $\gamma$ adopted by $SS+L_2$.\\
The top panels of Fig. \ref{FigIE} display the impulse response estimates
obtained by $SS+L_1$ (dashed line). The advantage of the new robust formulation is evident.
While $SS+L_1$ and $SS+L_2$ exhibit a similar performance under nominal conditions
(top left panel), $SS+L_1$ outperforms $SS+L_2$  in presence of outliers (top right panel),
returning an impulse response estimate much close to truth. 
The robustness of the $L_1$ loss w.r.t. large model deviations is due to the fact that it pushes some residuals to zero. 
It thus detects which measurements are more accurate,
essentially treating them as constraints during the fitting.\\
To further compare $SS+L_2$ and $SS+L_1$, Fig. \ref{FigIE2}
plots the fit (\ref{eq:fit}) returned by these two estimators as a function
of the regularization parameter $\gamma$ (with $\alpha$ constant, fixed to its estimate). 
The figure reveals that the adoption of the quadratic loss
makes it difficult to choose
the regularization parameter:
many values of $\gamma$ lead to poor estimates,
e.g. $\gamma \leq 1$ leads to outliers overfitting. On the other hand the fit profile 
associated with $SS+L_1$ is more stable, and is  
uniformly better than $SS+L_2$.\\ 
We have also tested the Vapnik loss formulation of the stable spline estimator.
The fit profile is displayed in Fig. \ref{FigIE2} 
(parameters $\alpha$ and $\epsilon$ are constant, set to their 
cross validated estimates) and is similar to the
one obtained by $SS+L_1$. Even though it requires estimation of the
additional parameter $\epsilon$, an advantage of the Vapnik loss
over the $L_1$ is its data compression capability: it detects the so called support vectors
which contain those measurements influencing the estimate, see \cite{Hastie90}
for details.\\
The last estimator tested embodies the information that the
data come from a relaxation system. This means that  
the impulse response is a completely monotonic function (e.g. see Section 4 in \cite{Cucker})
and, hence, its derivatives $f^{(\ell)}$ satisfy  $(-1)^{\ell} f^{(\ell)}(t) \geq 0$ for $t \geq 0, \ \ell=0,1,\ldots$.
Since $x=Ly$, in discrete-time
this information is (approximately) encoded in (\ref{genELQP})
by setting $a$ to the null vector and
\[
\begin{aligned}  
A^T =  \begin{bmatrix} -I_n  \\ D^1 \\ \ldots \\   (-1)^{k-1}D^{k}  \end{bmatrix}L,\; 
\end{aligned}
\]
with $k$ a sufficiently large integer and $D$ an $n \times n$ lower triangular Toeplitz matrix whose first column is $[1,-1,0,\ldots]^T$.
Let $SS+Rel$ denote the estimator (\ref{SS+L1}) complemented with 
the above constraints ($k=5$), with the parameter $\alpha$  set to the estimate 
used by $SS+L_1$. The corresponding fit is reported in Fig. \ref{FigIE2}: 
 $SS+Rel$ shows impressive performance for a wide range of regularization
parameter values. Interestingly, the model incorporating the complete monotonicity constraint
competes favorably with the overfit residuals even for very low values of $\gamma$.
This example illustrates the observation that the inclusion of additional model information
in the form of constraints often helps to regularize the estimation process while 
simultaneously improving the fit.

  \subsection{Computational efficiency: comparison of IPsolve, TFOCS and libSVM}\label{CVX}


In this section, we compare the computational efficiency of the IPsolve approach with those of 
two widely used publicly available packages, TFOCS~\cite{TFOCS} and libSVM~\cite{chang2011libsvm}. 
While in fields such as machine learning  it is often assumed that the number of unknowns is much larger than the data set size $m \gg n$, 
in system identification it is typically assumed that
$n \gg m$. This makes the per iteration complexity 
$O(n^2(m+ n))$ of our algorithm (see Corollary \ref{sysIDcor})
particularly appealing. We illustrate the advantages 
for the specific problem
\begin{equation}\label{SS+Vapnik}
\arg \min_{x} \|z-\Phi x\|_{\epsilon}  + \gamma x^T Q^{-1} x\;,
\end{equation}
where $\| \cdot \|_{\epsilon}$ denotes the Vapnik loss (Fig. \ref{fig:vapnik}), with noisy data $z$ generated from the  impulse response used in the introductory example (Fig. \ref{FigIE}, top panel). 
The objective is specified using parameters $\gamma=10$, $\alpha=0.5$ in (\ref{eq:TC}), and $\epsilon=0.1$. %

\subsubsection{TFOCS}

The TFOCS algorithm~\cite{TFOCS} is based on the proximal point algorithm, and can be applied to generic convex minimization problems.
The TFOCS software\footnote{\url{http://cvxr.com/tfocs}}  can handle a broader class of problems than 
IPsolve (i.e. problems that are not 
piecewise linear quadratic). The relevant standard form for TFOCS is the problem 
\begin{equation}
\label{eq:tfocsProbm}
\min \phi(x) = f(x) + \frac{\mu}{2}\|x-x_0\|^2 + h(\mathcal{A}(x) - b),
\end{equation}
where the proximity operator for both $f$ and $h$ can be efficiently computed. 
TFOCS combines dual smoothing techniques with optimal 
first-order methods~\cite{TFOCS,nest_orig,nesterov2005smooth} 
and is therefore capable of solving large-scale problems (much larger than those 
that can be solved with other general convex solvers, such as the MATLAB package CVX \cite{CVX2,CVX1}.).  
Like IPsolve, it is a general purpose software that can be used to solve~\eqref{SS+Vapnik}.

 \subsubsection{libSVM}
 
 libSVM is an optimization package\footnote{\url{https://www.csie.ntu.edu.tw/~cjlin/libsvm/}} aimed at 
 support vector problems, including problems of type~\ref{SS+Vapnik}. 
 It uses pre-compiled routines with several interfaces, including one for Matlab. libSVM is designed for 
 a broad range of support vector problems, including kernel machines; 
 for our problem of interest, the formulation available through libSVM is a slight modification of~\eqref{SS+Vapnik}:
 \begin{equation}\label{SS+Vapnik+mod}
\arg \min_{x,b} \|z-\Phi x-b\mathbf{1}\|_{\epsilon}  + \gamma x^T Q^{-1} x.\;
\end{equation}
The inclusion of intercept $b$ is not optional for libSVM; we therefore 
compared libSVM with IPsolve on~\eqref{SS+Vapnik+mod}. libSVM solves the {\it dual} problem to~\eqref{SS+Vapnik+mod}:
 \begin{equation}
 \label{eq:dualSVM}
\begin{aligned}
 \min_{\alpha\in \frac{1}{\gamma}\mathbb{B}_{\infty}} 
& \frac{1}{2}\alpha^T(Q^{T/2}\Phi^T\Phi Q^{1/2})\alpha 
 + \epsilon \|\alpha\|_1 + \langle z, \alpha \rangle, \quad \mbox{ s.t.} \quad 1^T\alpha = 0. 
\end{aligned}
 \end{equation}
This is a quadratic problem, and libSVM solves it using a specialized decomposition approach. 
By focusing on the dual, libSVM is able to handle linear and nonlinear SVM 
and SVR; it has been widely applied in practice. 

In the standard system identification context, we have $n<<m$; and as shown in 
Corollary~\ref{sysIDcor}, the number of arithmetic operations required to implement 
each iteration is $O(n^2(n+m))$. In contrast, a naive approach for solving~\eqref{eq:dualSVM}
requires $O(m^3)$ operations. While the approach of~\cite{chang2011libsvm} is far from naive,
it is optimized for kernel machines, where one must restrict all computation to the $m$-dimensional 
dual (since the primal dimension may be infinite); in contrast IPsolve exploits the structure of the problem, performing most computations for~\eqref{SS+Vapnik} in an $n$-dimensional space.

\subsubsection{Experimental setup and results}
We compare all three approaches for problem~\eqref{SS+Vapnik} using 
$n \in \{100, 150, 200, 250, 300\}$ and $m \in \{1000, 2000, 5000, 10000, 20000\}$.
The scales of $(m,n)$ are chosen to reflect the common system identification context, where $n<<m$. 
We run each algorithm until their available optimality criteria fall below $\epsilon = 10^{-6}$; 
The precise criteria for the three algorithms are as follows. 

IPsolve uses a relative magnitude of the Karush-Kuhn-Tucker system;  
it terminates when 
\(
\frac{\|F_{\text{current}}\|}{\|F_{\text{initial}}\|} < \epsilon,
\) for $F = F_0$ in~\eqref{F mu}.

TFOCS allows the user to select one of several stopping criterias; some are based on optimality of problem~\eqref{eq:tfocsProbm}; but there is also a relative criteria based on iterate convergence, 
\(
\frac{\|x_{k+1} - x_k\|}{\max(1, \|x_{k+1}\|)} < \epsilon
\)
 that allows the algorithm to terminate early. This is the criteria we selected; internal optimality criteria required a far larger number of iterations, during which the objective value did not change significantly. To optimize performance of TFOCS, we experimented with choice of first-order solvers, but found the default algorithm (Auslender \& Teboulle's single-projection method) to be the best. We also tuned the `restart' option to restart the step-length computation every 1000 iterations; as this improved performance of TFOCS, as recommended by the authors.

libSVM convergence criteria for 
the SVR problem, as explained in~\cite{chang2011libsvm},  is based on the iterate $\alpha$ satisfying 
a KKT criteria for~\eqref{eq:dualSVM} within $\epsilon$ (in the infinity norm).

In summary, we have two similar optimization problems,~\eqref{SS+Vapnik} 
and~\eqref{SS+Vapnik+mod}, and three sets of convergence criteria. To fairly compare
the algorithms, we run IPsolve against TFOCS on problem~\eqref{SS+Vapnik}, and 
IPsolve against libSVM on problem~\eqref{SS+Vapnik+mod}. In each case, 
we tabulate both timing results, and also show an `accuracy' heuristic, which is the 
signed {\it relative objective difference} (ROD): 
\[
\mbox{ROD}(*):= (f_{\mbox{(*)}} - f_{\mbox{IPsolve}})/f_{\mbox{IPsolve}}
\]
A positive ROD indicates IPsolve found the lower objective value; a relative scale 
is chosen because we consider a range of problem sizes. 

Results of the numerical study are shown in Tables~\ref{TabTFOCS} and~\ref{TabLIBSVM}. IPsolve 
gets uniformly better objective values for all experiments, and performs faster 
that TFOCS for all problem sizes.  Notably, TFOCS is very 
accurate at the settings we compared, with all ROD values less than $10^{-5}$. 

libSVM is more competitive in its timing, but also less accurate, with some 
ROD values exceed $10^{-2}$, i.e. libSVM primal values are more than 1\% larger than those of IPsolve
on some of the problems. Overall, IPsolve converges faster for approximately half of the problems; it especially has an advantage for large $m$ and small $n$ as expected. It should be noted that libSVM has strange behavior for the $n=300$ case; it converges very quickly, but the solutions are less accurate than for other problems.

\begin{table}
\begin{center}
\footnotesize{
\begin{tabular}{cccccc}\hline
     \diaghead{\theadfont}{\theadfont m}{\theadfont n}       & $1K$ & $2K$ & $5K$ & $10K$ & $20K$\\ 
     \\
     \hline
%
%
$100$  &      { 0.376 } &  0.730  &  1.333  &2.666  &5.365\\
$150$ &     0.426  &  0.886  & 2.782 &  4.225 &  7.862 \\
 $200$  &      0.464  &  1.149  &  3.040  & 4.796 & 10.105\\
$250$ &        0.464   & 1.273 &   3.541 &  6.176  &11.463\\
$300$  &       0.715   & 1.633  &  3.966  &  7.418  &  13.773\\ 
 \hline
\phantom{|}
\end{tabular} \;\;\begin{tabular}{cccccc}\hline
     \diaghead{\theadfont}{\theadfont m}{\theadfont n}       & $1K$ & $2K$ & $5K$ & $10K$ & $20K$\\ 
     \\
     \hline
%
%
$100$  &     3.320 &   5.292  &  5.815  & 10.896  & 20.483\\
$150$ &         4.882 &   4.196  &  6.800  & 11.194 &  26.349 \\
 $200$  &         3.545   & 3.985  &  9.282  & 10.710  & 32.644\\
$250$ &        3.840  &  5.167  &  8.722  & 14.787  & 30.289 \\
$300$  &        3.256  &  4.471  &  6.787  & 11.186  & 17.403\\
  \hline
\phantom{|}
\end{tabular}}\\
{ ROD(TFOCS):}\quad$10^{-5}$\begin{tabular}{cccccc}\hline
     \diaghead{\theadfont}{\theadfont m}{\theadfont n}       & $1K$ & $2K$ & $5K$ & $10K$ & $20K$\\ 
     \\
     \hline
%
%
$100$  &     0.0418 &   0.0046  &  0.0024  & 0.0044  & 0.0022\\
$150$ &         0.0179 &   0.0080  &  0.0015  & 0.0068 &  0.0260 \\
 $200$  &         0.0175   & 0.0135  &  0.0053  & 0.0055  & 0.0399\\
$250$ &        0.0074  &  0.0055  &  0.0028  & 0.0208  & 0.0100 \\
$300$  &        0.1423  &  0.1424  &  0.1756  & 0.0982  & 0.1748\\
  \hline
\phantom{|}
\end{tabular}\\
\end{center}
\caption{Top tables: CPU time (seconds) taken by IPsolve (left) and TFOCS (right) to solve~\eqref{SS+Vapnik}
as a function of the dimension $n$ of $x$ and of the data set size $m$.
Bottom table: relative (signed) objective difference of the solutions. IPsolve {\it always} gets the lower objective 
value; and it is uniformly faster for all problems tested. Note that TFOCS is very accurate. }
\label{TabTFOCS}
\end{table} 

\begin{table}
\begin{center}
\footnotesize{
\begin{tabular}{cccccc}\hline
     \diaghead{\theadfont}{\theadfont m}{\theadfont n}       & $1K$ & $2K$ & $5K$ & $10K$ & $20K$\\ 
     \\
     \hline
%
%
$100$  &      { 0.293 } &  0.629  &  1.303  &2.427  &5.011\\
$150$ &     0.421  &  0.791  & 2.061 &  4.197 &  7.522 \\
 $200$  &      0.504  &  1.111  &  2.791  & 4.827 & 9.040\\
$250$ &        0.621   & 1.148 &   3.263 &  6.086  &10.460\\
$300$  &       0.719   & 1.271  &  3.865  &  7.314  &  13.077\\ 
 \hline
\phantom{|}
\end{tabular} \;\;\begin{tabular}{cccccc}\hline
     \diaghead{\theadfont}{\theadfont m}{\theadfont n}       & $1K$ & $2K$ & $5K$ & $10K$ & $20K$\\ 
     \\
     \hline
%
%
$100$  &     0.764 &   0.343  &  2.305  & 8.528  & 32.944\\
$150$ &         0.124 &   0.485  &  2.930  & 11.628 &  46.038 \\
 $200$  &         0.153   & 0.586  &  3.691  & 14.909  & 59.223\\
$250$ &        0.161  &  0.667  &  4.348  & 17.885  & 72.330 \\
$300$  &        0.052  &  0.107  &  0.450  & 1.515 & 5.330\\
  \hline
\phantom{|}
\end{tabular}}\\
rod(libSVM):\quad  $10^{-3}$\begin{tabular}{cccccc}\hline
     \diaghead{\theadfont}{\theadfont m}{\theadfont n}       & $1K$ & $2K$ & $5K$ & $10K$ & $20K$\\ 
     \\
     \hline
%
%
$100$  &     1.8080 &   0.2988  &  0.0284  & 0.1225  & 0.0037\\
$150$ &         1.4702 &   0.1488  &  0.0241  & 0.0997 &  0.0049 \\
 $200$  &         3.5449   & 0.1984  &  0.0172  & 0.1130  & 0.0004\\
$250$ &        1.5892  &  0.1723  &  0.0141  & 0.1643  & 0.0080 \\
$300$  &        5.4052  &  21.9701  &  3.8528  & 0.4366  & 1.2052\\
  \hline
\phantom{|}
\end{tabular}\\
\end{center}
\caption{ Top tables: CPU time (seconds) taken by IPsolve (left) and libSVM (right) to solve~\eqref{SS+Vapnik+mod}
as a function of the dimension $n$ of $x$ and of the data set size $m$.
Bottom table: relative (signed) objective difference of the solutions. Note that IPsolve {\it always} gets the lower objective 
value, and it is faster for the majority of the problems.
 Note also that libSVM is a lot less accurate than TFOCS (accuracy measured by IPsolve objective). 
 }
\label{TabLIBSVM}
\end{table}

\subsection{Monte Carlo study in the presence of outliers}\label{MC}

We now  consider a Monte Carlo study of 1000 sample runs. 
For each run, a random single-input single-output~(SISO)~ 
continuous time
model of order 30 is generated and then sampled at three times its bandwidth using Matlab commands:
\begin{Verbatim}[fontseries = b, fontsize = \small]
	m=rss(30); b=bandwidth(m); f = b*3*2*pi; 
	md=c2d(m,1/f,'zoh'); md.d = 0;
\end{Verbatim}
We accept only models with all poles in a disk of radius $0.95$ in the complex plane. 
The model \verb{md{, initially at rest, is given a Gaussian white noise input 
with unit variance filtered by a randomly generated model obtained by 
the same process already described. The input delay is always equal to 1.
We then generate 1000 measurements contaminated by outliers, 
and use them to reconstruct the impulse response. 
The measurement errors are a mixture of 
two normals given by
$e_i \sim 0.7 {\bf{N}}(0,\sigma^2) + 0.3  {\bf{N}}(0,100\sigma^2)$,
with $\sigma^2$ equal to the variance of the noiseless
output divided by 100. Thus, with probability
0.3, a measurement becomes an `outlier', since the corresponding simulated error 
has standard deviation $10\sigma$.
\\ 
We compare the performance of five different estimators over the 1000 runs of the simulation. 
Each estimator uses either quadratic or 1-norm loss for data fidelity, 
and all formulations treat the system input delay and initial conditions as known.  
The estimators are enumerated below. 

\begin{figure*}
  \begin{center}
   \begin{tabular}{cc}
\hspace{-.4in}
 { \includegraphics[scale=0.36]{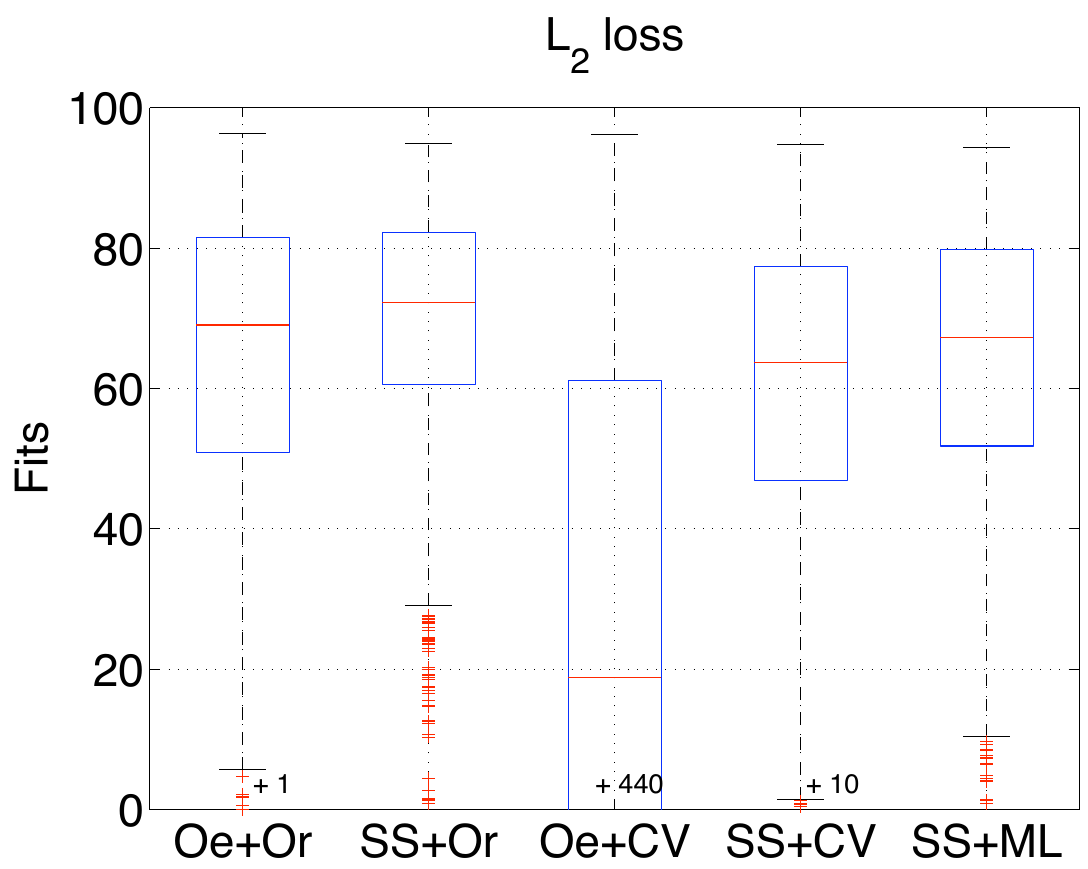}} 
\hspace{.0in}
{\includegraphics[scale=0.36]{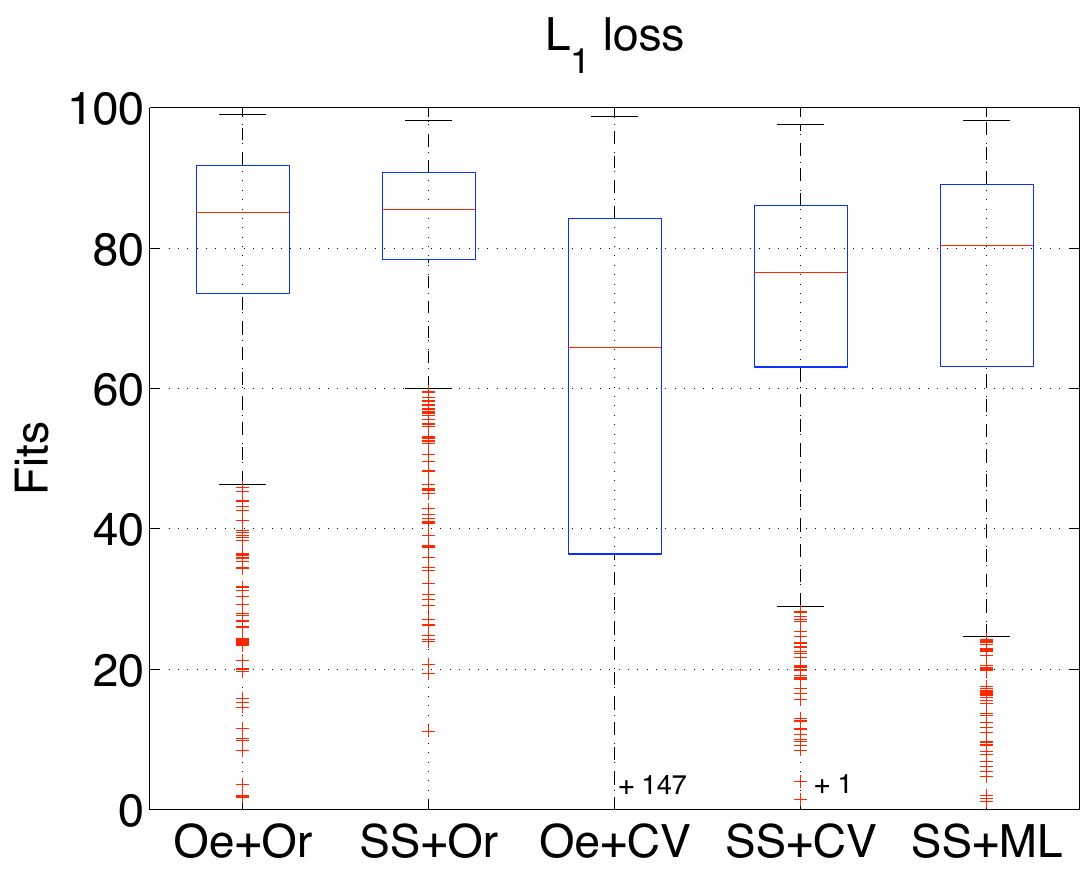}}
    \end{tabular}
 \caption{Monte Carlo study (subsection \ref{MC}). 
 Boxplot of the 1000 percentage fit measures (\ref{eq:fit}) obtained by PEM and stable spline estimators
 equipped with the $L_2$ loss (left panel) and the $L_1$ loss (right panel).} 
    \label{FigOE}
     \end{center}
\end{figure*}

\begin{table}
\begin{center}
\begin{tabular}{cccccc}\hline
            & \textit{Oe+Or} & \textit{SS+Or} & \textit{Oe+CV} & \textit{SS+CV} & \textit{SS+ML} \\\hline
$L_2$ loss  & 64.31 &  68.9  & -303.2 & 59.7 & 63.1\\
$L_1$ loss   & 79.9 & 82.3 & -73.1 & 71.8 & 72.8 \\
 \hline
\phantom{|}
\end{tabular}
\end{center}
\caption{Monte Carlo study (subsection \ref{MC}).
Average fit achieved by the PEM and stable spline estimators 
using the $L_2$ and the $L_1$ loss.}
\label{Tab1}
\end{table} 

\noindent
{\it Oe+Or} is the classical PEM approach (MATLAB command \texttt{oe}), 
and we compare
the quadratic loss (\ref{eq:5}) with the $1$-norm loss $ V(x)=\sum^m_{t=1}\left|y(t)-G(q,x)u(t) \right|$. 
Candidate models are rational  transfer functions (\ref{eq:BB}) 
with polynomials $B$ and $C$ of the same order. 
The estimator is not implementable in practice, since it uses an oracle 
which selects the model order maximizing the percentage fit measure (\ref{eq:fit}),
and provides the {\it best achievable} PEM performance.    

\noindent
{\it Oe+CV} is an implementable analogue of {\it Oe+Or}  that uses cross-validation to 
estimate model order. Data are split into training and validation sets of equal size. 
For every model order ranging from 1 to 30, a model 
is trained using command \verb{oe{ on  the training set.
We choose the order that minimizes the sum of squared prediction errors 
on the validation set, obtained by using the command \verb{predict{ with null initial conditions.
Once the order is found, the final model is computed by \verb{oe{ using all measurements 
in training and validation sets. 

\noindent
{\it SS+Or}  is the stable spline estimator
using  (\ref{eq:TC}). Again, we compute the fit using both  
the quadratic loss (\ref{eq:MV}) and the
1-norm loss (\ref{SS+L1}) for the purpose of comparison.
The number of estimated impulse response coefficients 
is 200, i.e. $\dim(x)=200$.
This estimator also uses an oracle which gives values for the hyperparameters 
 $\alpha$ and $\gamma$ that maximize the percentage fit measure (\ref{eq:fit}).
As is the case for \textit{Oe+Or}, this method is not implementable, but provides 
a baseline for the {\it best possible} performance of a stable spline estimator.

\noindent
\blue{
 {\it SS+ML} is an implementable analogue of {\it SS+Or} 
that uses marginal likelihood maximization to estimate hyperparameters.
{\bf For the quadratic loss},  
we estimate $\sigma^2$ (the noise variance) by fitting a low-bias FIR model of order $p$, 
(see e.g. in \cite{Goodwin1992} for details), and then set
\begin{equation}\label{eq:sigmaes}
\hat{\sigma}^2 = \sum^m_{t=1}\left(y(t)-\hat G(q)u(t) \right)^{2}/(m-p),
\end{equation}
where $\hat G$ is the $p$th-order FIR obtained by least-squares. We 
estimate $\alpha$ and $\gamma$ 
by maximizing the {\it marginal likelihood}
\begin{equation}\label{eq:marglik}
(\hat \lambda, \hat \alpha) 
=
\arg \min_{\lambda,\alpha} \left\{z^T \Sigma^{-1} z + \log \det (\Sigma)\right\}, \quad \Sigma := \lambda \Phi Q \Phi^T + \hat{\sigma}^2 I_m \;.
\end{equation}
%
Then let $\hat Q = Q(\hat \alpha)$ and $\hat \Sigma = \Sigma(\hat \lambda, \hat \alpha)$ 
be the estimates of $Q$ and $\Sigma$ 
with $\lambda = \hat \lambda$ and $\alpha = \hat \alpha$. From
\eqref{eq:MV}, the final impulse response estimate becomes 
$\hat x = \hat \lambda \hat Q \Phi^T \hat \Sigma^{-1} z$. 
{\bf For the 1-norm loss},
we model components of $E$ in (\ref{MatrixMod}) 
as independent Laplacian random variables with variance $\sigma^2$.
For known hyperparameters, the negative log posterior of $(x|z)$ is 
$$
\sqrt{\frac{2}{\sigma^2}} \|z-\Phi x\|_1 + \frac{1}{2 \lambda}  x^T Q^{-1} x
$$
with  constant terms omitted.
Then (\ref{SS+L1}) is the MAP estimator of $x$ given $z$ if
\begin{equation}\label{L1gamma}
\gamma = \frac{\sigma^2}{2 \sqrt{2} \lambda}\ .
\end{equation}
We take 
$\lambda$ and $\alpha$ to be the same estimates as obtained under
Gaussian noise assumptions (i.e. optimizing (\ref{eq:marglik})), then use
(\ref{L1gamma}) and  (\ref{eq:TC}) to obtain $\gamma$ and $Q$, respectively. \\
}

{\it SS+CV} is nearly identical to \textit{SS+ML}, but with 
hyperparameters estimated by cross-validation. 
Data are split into a training and validation 
set of equal size and the best values of $\gamma$ and $\alpha$
are found over a two dimensional grid. 
Specifically, the $\alpha$-$\gamma$ grid
is $\texttt{A}\times\texttt{B}$ , where $A$ and $B$ are given by the
MATLAB commands:

\begin{Verbatim}[fontseries = b, fontsize = \small]
	 A=[0.01 0.05:0.05:0.95 0.99]  
	 B=logspace(log10(g/100),log10(g*100),50)
\end{Verbatim}
with \texttt{g} taken to be the value of $\gamma$ used in \textit{SS+ML}.
%
%
The plots in Fig. \ref{FigOE} show Matlab
boxplots of the 1000 percentage fit measures (\ref{eq:fit}) obtained by the five estimators.
The rectangle contains the inter-quartile range ($25-75\%$ percentiles)  
of the fits, with median shown with a red line. 
The ``whiskers" outside the rectangle display the upper and
lower bounds of all the numbers,
not counting what are deemed outliers,
plotted separately as ``+". Table \ref{Tab1}
also reports the average percentage fit values.\\
The left panel of Fig. \ref{FigOE} shows the fits achieved
by the quadratic estimators.
Oracle-based procedures highlight the advantage of the stable spline estimators:
\textit{SS+Or} shows better performance than 
$\textit{Oe+Or}$. The performance gap increases in implementable 
estimators, when hyperparameters are learned from data (as necessary in practical situations).
 \textit{SS+CV} and \textit{SS+ML}
have  similar performance, and both are much better than \textit{Oe+CV}.\\
The right panel of Fig. \ref{FigOE} displays the fits achieved
by all five estimators when using the 1-norm data fidelity loss function.
These estimators are more robust against outliers, and all
fits improve significantly. 
Furthermore, as in the previous case, performance of stable spline estimators 
is superior to that of the classical system identification procedures.

\subsection{Assessment of cerebral hemodynamics using magnetic resonance imaging} \label{MRI}

The quantitative assessment
of the cerebral blood flow is essential to the understanding of brain function. 
An important technique is bolus-tracking magnetic resonance imaging (MRI), which relies on 
established principles for tracer kinetics of nondiffusible tracers \cite{MRI2}.
These principles allow for the quantification of cerebral hemodynamics by solving a linear system
identification problem. The system output is the measured tracer concentration within a given tissue volume of interest,
while the system input is the measured arterial function. 
The impulse response is proportional to the so called {\it tissue residue function}, and is known to be positive and unimodal. 
It carries fundamental information on the system under study, e.g. the cerebral blood flow is given by its maximum value. However, 
impulse response estimation is especially difficult for this problem: even if the noise
can be reasonably modeled as Gaussian, the problem is often
ill-conditioned and only a few noisy output samples are available \cite{MRI4}.\\
\noindent
We consider realistic simulation studies, using four different types of estimators 
all based on the quadratic loss. 
The first two rely on the classical PEM paradigm. They are 
$\textit{Oe+Or}$, described in the previous subsection,
with the maximum allowed model order equal to 10,
and $\textit{Oe+AICc}$, which uses the AICc criterion \cite{Hurvich} 
for model complexity selection. 
$\textit{SS+ML}$ uses the stable spline kernel (\ref{eq:SS}), 
estimates hyperparameters via marginal likelihood optimization, 
and then uses  (\ref{eq:MV}) to find the final impulse response, 
where $x \in \mB{R}^{100}$.
The last estimator $\textit{SS+ML+um}$ 
also estimates hyperparameters via marginal likelihood optimization, 
and then incorporates nonnegativity and unimodality information.
Specifically, we minimize objective (\ref{eq:MV}) (with hyperparameter estimates
identical to those used by $\textit{SS+ML}$), 
subject to  {\it inequality constraints} that impose unimodality: 
\begin{equation} 
  \label{Umconstraints}
    D_1^k x_1^k  \geq 0, \quad 
    D_{k+1}^{n} x_{k+1}^{n}  \leq 0, \quad 
    x  \geq 0\ .
\end{equation}
\begin{figure*}[h!]
  \begin{center}
   \begin{tabular}{cc}
\hspace{-.34in}
 { \includegraphics[scale=0.37]{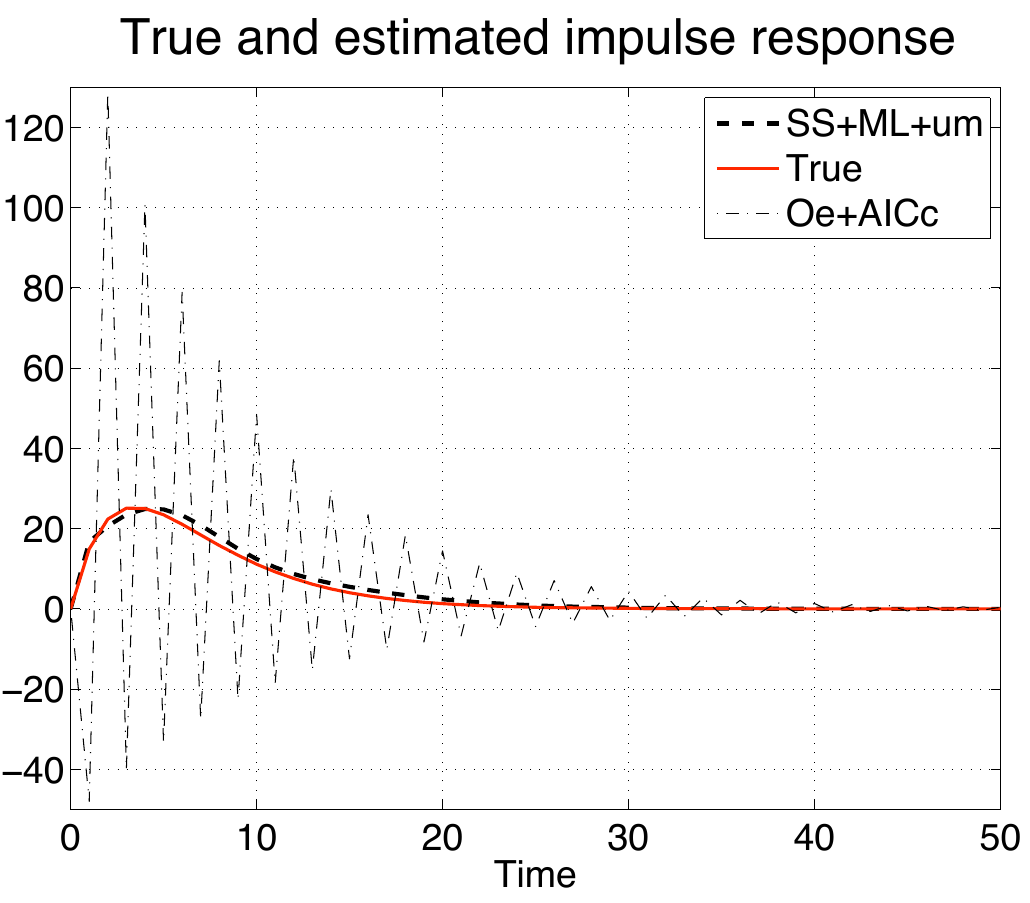}} 
\hspace{.1in}
{\includegraphics[scale=0.37]{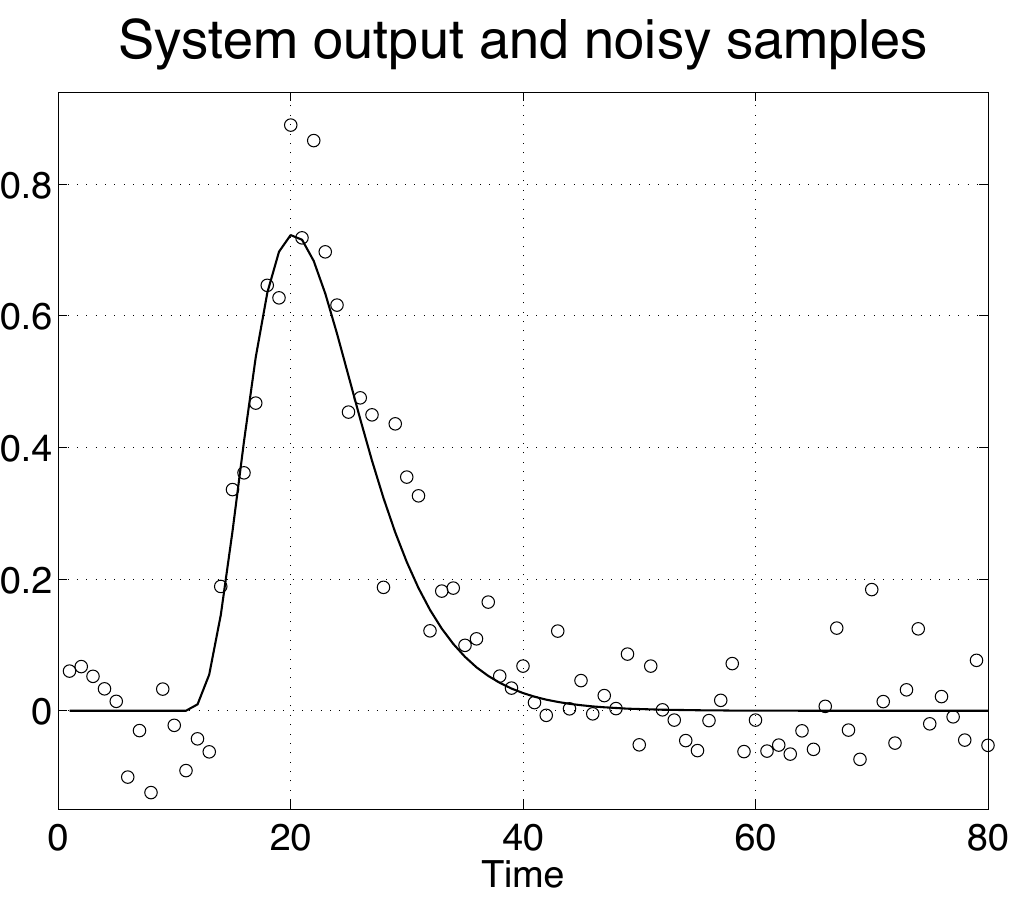}}
    \end{tabular}
    \caption{Cerebral hemodynamics. 
{\it (Left)}: true impulse response (solid), 
PEM estimate with AICc to select the model order (dashdot) and 
the stable spline estimate incorporating unimodality constraints (dashed). 
{\it (Right)}: Noiseless output (solid) and
the measurements ($\circ$).} \label{FigMRI1}
     \end{center}
\end{figure*}

\begin{figure*}[h!]
  \begin{center}
   \begin{tabular}{cc}
\hspace{-.5in}
 { \includegraphics[scale=0.37]{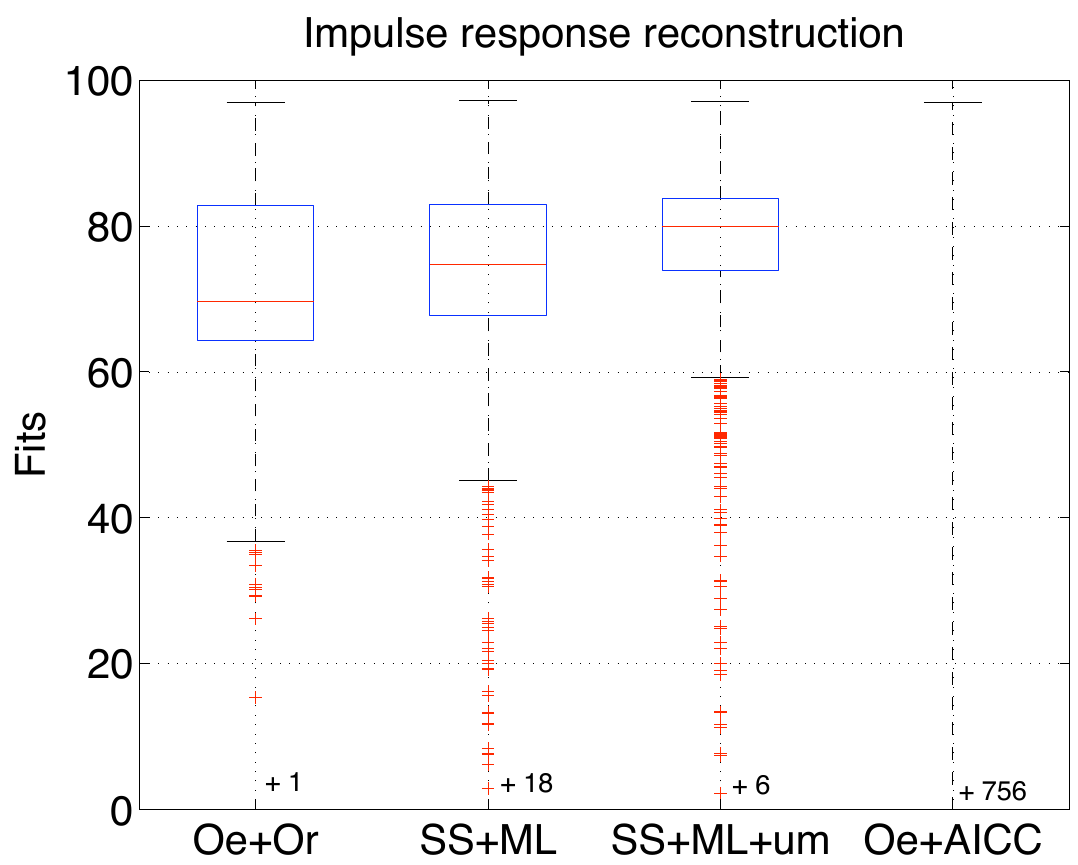}} 
\hspace{-.1in}
{\includegraphics[scale=0.37]{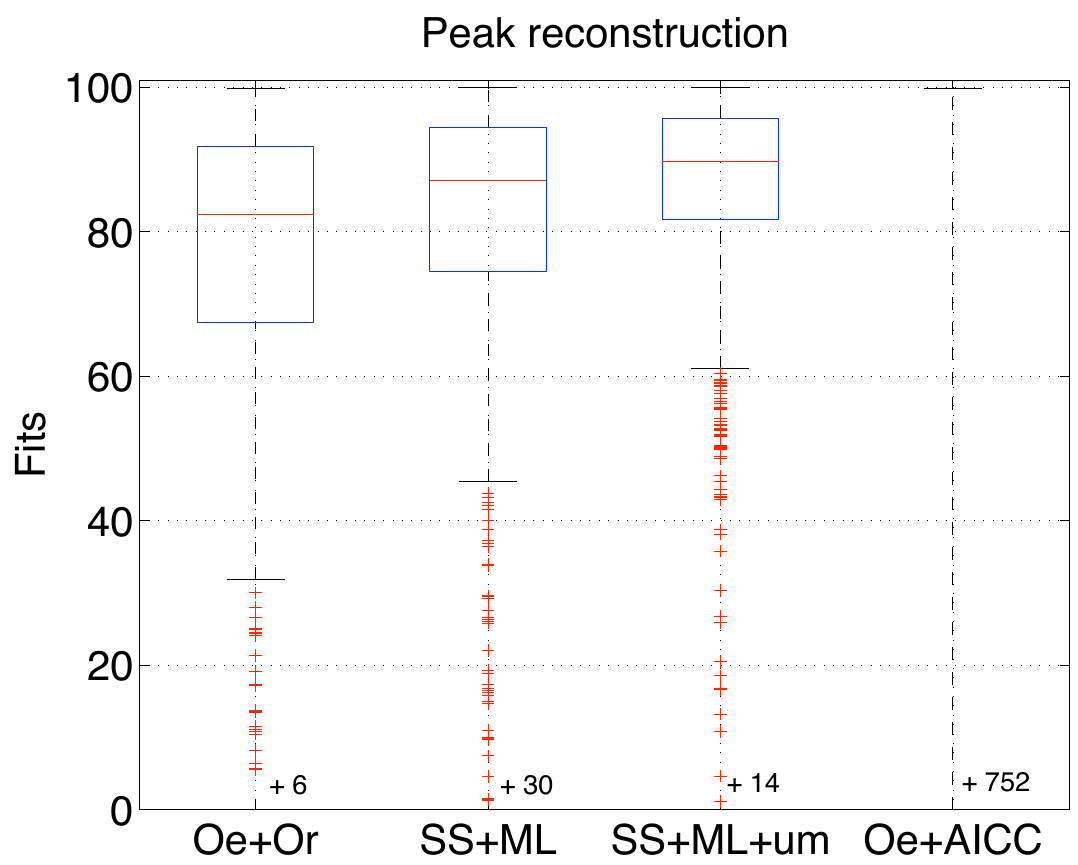}}
    \end{tabular}
\caption{Assessment of
cerebral hemodynamics. 
Boxplot of the 1000 
percentage fit measures (\ref{eq:fit}) for the impulse response reconstruction (left) and peak 
reconstruction (right) obtained by PEM equipped with an oracle and with AICc
for model order selection, by the stable spline estimator and by the 
stable spline estimator incorporating unimodality constraints.} \label{FigMRI2}
     \end{center}
\end{figure*}
Here, $D \in \mB{R}^{n \times n}$ is the discrete derivative operator, 
i.e. a lower triangular Toeplitz matrix with first column $[1,-1,0,\ldots]^T$, 
$D_1^k$ and $D_{k+1}^{n}$ contain, respectively, 
the first $k$ and the last $n-k$ rows of $D$,
and analogously for $x_1^k$ and $x_{k+1}^{n}$.
In terms of (\ref{genELQP}), this is specified setting $a$ to the null vector and
\begin{equation}
\label{eq:mono}
\begin{aligned}
A = L^T\begin{bmatrix}  -(D_1^k)^T & (D^n_{k+1})^T  &  -I_n  \end{bmatrix}.\; 
\end{aligned}
\end{equation}
We solve the problem for each $k$, obtaining a set of solutions $\hat x(k)$, 
and then select 
the best $\overline k$ for~\eqref{eq:MV} by setting 
$\overline k = \arg\min_k   \|z-\Phi \hat x(k)\|^2  + \gamma (\hat x(k))^T Q^{-1} \hat x(k)\;$,
with the final estimator given by $\hat x(\overline k)$, the best unimodal estimator.
We begin by considering the simulation described in 
\cite{MRI4}. The system input is the typical arterial function $u(t)=(t-10)^3e^{-\frac{2t}{3}}$
if $t>10$ and zero otherwise,
while the impulse response is the dispersed exponential displayed
in Fig. \ref{FigMRI1} (left panel, solid line). This response is to be reconstructed
from the 80 noisy output samples reported in Fig. \ref{FigMRI1} (right panel).
These measurements are generated as in subsection II.A of \cite{MRI4}, using
parameters typical of a normal subject with a signal to noise ratio
equal to 20 and discretizing the problem using unit 
sampling instants.\\ 
The left panel of Fig. \ref{FigMRI1} shows the estimate by $\textit{Oe+AICc}$ (dashdot).  
The reconstructed profile is far from the truth and contains 
many non-physiological oscillations: the asymptotic theory underlying AICc
does not compensate for ill-conditioning.
The same panel shows the $\textit{SS+ML+um}$ estimate (dashed line).
This estimate is very close to truth, outlining the importance of 
regularization and the inclusion of addition information in the form of
constraints when handling these data poor situations.
This result
is confirmed by a Monte Carlo study of 1000 runs 
where independent noise realizations are generated at every run. 
Fig. \ref{FigMRI2} shows the Matlab boxplots of the 1000 percentage fit measures (\ref{eq:fit}) 
achieved by the four estimators in the reconstruction of the
impulse response (left panel) and of its peak (right panel).
Most of the time $\textit{Oe+AICc}$ returns negative fits,
while the outcomes from $\textit{Oe+Or}$ and $\textit{SS+ML}$
are similar with good performance (keep in mind $\textit{Oe+Or}$ is not implementable in practice).
Finally, notice that $\textit{SS+ML+um}$ has the 
best performance vis-\`a-vis the
percentage fit measures (\ref{eq:fit}), see also Table \ref{Tab2}
for the average fits.

\begin{table}[h!]
\begin{center}
\begin{tabular}{ccccc}\hline
            & \textit{Oe+Or} & \textit{SS+ML} & \textit{SS+ML+um} & \textit{Oe+AICc}  \\\hline
Imp. response  & 71.9 &  69.8  & 76.3 & -3497.1 \\
Peak  & 77.1 & 76.2 & 84.3 & -26.3  \\
 \hline
\phantom{|}
\end{tabular}
\end{center}
\caption{Average percentage fit (\ref{eq:fit}) achieved by the PEM and stable spline estimators
relative to impulse response and peak reconstruction.}
\label{Tab2}
\end{table} 

\blue{
  \subsection{Stable and sparse estimation}\label{SparseStable}
  A  multiple input single output (MISO) system can be written as 
\begin{equation}\label{MatrixMod}
  z=\sum_{j=1}^p \Phi^j x^j + E,   
\end{equation}
where each vector $x^j$ contains the coefficients of the $j$-th impulse response.
This problem is challenging since the number of unknowns can be much larger than the number
of output measurements. One strategy is to use sparse regularization on $x^j$ to simultaneously perform variable selection and estimation. 
This problem arises in dynamic networks, where sparsity is used to detect structural connectivity.\\
To design a sparsity promoting stable PLQ estimator, take $Q$ be the stable spline kernel in (\ref{eq:TC}) 
and let let $y^j = L^{-1}x^j, \ Q = LL^T$. Then solve the problem 
\begin{equation}
\label{SysIdObjectiveSparseStable} 
\min_y 
\left\|z - \sum_{j=1}^p \Phi^j L y^j\right\|^2+  \sum_{j=1}^p \gamma \|y^j\|_1\;.
\end{equation}
}

\blue{
\subsubsection{A Monte Carlo study} A simple numerical study is used to test 
(\ref{SysIdObjectiveSparseStable}). Output data are generated as described in Section~\ref{MC},
but no outliers are introduced. In the MISO, $x^1$ is the impulse response 
plotted in the top panels of Fig. \ref{FigIE} while, for $j=2,\ldots,10$, the remaining $x^j$ are null. 
Each $x^j$ contains 100 impulse response coefficients. 
Thus, the estimator seeks to reconstruct 1000 unknowns from 400 measurements.
The regularization parameter $\gamma$ and the kernel parameter $\alpha$ are obtained using the 
cross validation strategy described in the Section~\ref{MC}.
At each Monte Carlo run we compute the fit (\ref{eq:fit}) related just to $x^1$. We also compute the fit
obtained by 
\begin{equation}
\label{SysIdObjectiveSparseStable2} 
\min_y 
\left\|z - \Phi^1 L y^1\right\|^2+  \gamma \|y^1\|^2\;,
\end{equation}
which corresponds to an oracle classical estimator that knows only $x^1$ may be different from zero.
After 100 runs the average fits of (\ref{SysIdObjectiveSparseStable}) and
 (\ref{SysIdObjectiveSparseStable2}) were {\bf $91.8$} and {\bf $93.4$}, that is, there was little loss in 
 fit relative to an oracle estimator.  
}

\subsubsection{A comparison with FISTA}
\blue{
The problem~\eqref{SysIdObjectiveSparseStable} is the sum of a smooth and simple function, 
and can therefore be optimized by primal-only first-order methods. 
The Fast Iterative Shrinkage Algorithm (FISTA)~\cite{Beck2009} is an {\it optimal} first-order method, 
i.e. it can obtain a guaranteed rate of convergence of $O(\frac{1}{k^2})$, matching the best complexity
of first-order methods for convex problems (with faster rates under stronger assumptions). 
However, the rate of convergence also depends on the condition number of the problem. \\
To compare IPsolve with this optimal first-order method in the high-dimensional system identification setting, 
we used the MISO problem to generate three scenarios: $400$ measurements vs. $1000, 5000$, and $10000$
unknown impulse responses. We solved~\eqref{SysIdObjectiveSparseStable} using IPsolve, and then ran FISTA to see how long it would take it to obtain the same function value. \\
The condition number of $L$ is $3.5 \times 10^5$ in each experiment. The condition numbers of 
$A = \Phi LL^T\Phi^T$ are given in the table. 
\begin{table}[h!]
\begin{center}
\begin{tabular}{cccccc}\hline
            & \textit{IP time} & \textit{IP iters}&  \textit{FISTA time} & \textit{FISTA iters} & \textit{Cond $A$}  \\\hline
400 x 1000  & 0.89 & 15 & 1.8  & 635 & $3.0\times 10^7$ \\
400 x 5000  & 1.96 & 16 &20 & 1204 & $1.27\times 10^4$  \\
400 x 10000  & 3.8 & 18&49.8 & 1458 & $2.83 \times 10^3$ \\
 \hline
\phantom{|}
\end{tabular}
\end{center}
\caption{Numerical comparison of IPsolve with FISTA. For each problem, FISTA is run until it hits the 
objective function achieved by IPsolve.}
\label{Tab3}
\end{table} 
When $n >> m$, IPsolve uses the Woodbury-Sherman-Morisson formula to form and invert an $m\times m$
matrix at each iteration. Therefore the dominant costs of each iteration are $O(m^2 (m+n))$, linear in $n$. 
Each iteration of FISTA is dominated by the computation of the gradient, which is $2mn$. On the other hand, 
the number of iterations required by FISTA is far less predictable than iterations required by IPsolve. 
The key difference between IPsolve and first-order methods are that its complexity per iteration 
is cubic in $\min(m,n)$, while that of FISTA is quadratic. However, in regimes where $m<<n$ or $n>>m$, 
and the systems may be ill-conditioned, IPsolve is competitive. 
Finally, the reader should keep in mind that~\eqref{SysIdObjectiveSparseStable} has a special form, 
and more general PLQ estimators cannot be solved with primal-only methods such as FISTA, but can 
still be solved using IPsolve.  
}

\section{Conclusions}\label{Conclusions}

This paper extends stable spline estimators to allow general modeling of 
misfit measures, regularizers, and constraints.
Quadratic losses and regularizers can now be replaced by 
general PLQ functions.  
Furthermore, affine inequality constraints
on the unknown impulse response can also be incorporated, 
providing a simple mechanism for the inclusion of information on domain restriction, 
monotonicity, and unimodality of the signal.
This
can have a profound impact on the quality of the recovery and can
significantly improve the fit   
when a regularizing prior is required for identification (as illustrated in Fig. \ref{FigIE2}).\\
\blue{The new framework allows the user to formulate and explore new system identification procedures, 
balancing robustness against outliers, introducing sparsity promoting priors, 
or introducing additional information by means of affine inequality constraints.
If the resulting nonparametric model is to be used for control purposes,
our estimates can be projected onto 
suitable low-order approximations using e.g. the approaches described in \cite{Benner2015}}.\\
All of the extensions have been implemented in the open source package IPsolve. 
Numerical comparisons in Section~\ref{CVX} showed that  
\blue{
IPsolve is well-suited to ill-conditioned problems that arise in system identification, 
and outperforms competing alternatives, including TFOCS and libSVM 
when $n<<m$ and first-order methods such as FISTA when $n>>m$.}\\
Multiple examples illustrate the power of the new framework in comparison to classical 
approaches. An important example comprises impulse responses known
to be positively or completely monotonic, in the presence of outliers. 
Classic approaches (using PEM and rational transfer function models \eqref{eq:BB})
solve a non-convex, possibly non-differentiable high-dimensional 
inequality constrained optimization problem ($\dim(x)$ specifies domain complexity)
{\it for each postulated model structure}. Model order selection,   
a delicate issue even in the unconstrained case, becomes, {\it{a fortiori}}, even harder.
These drawbacks have greatly limited the use of algorithms for inequality constrained/non-smooth
linear system identification. \\
In contrast,  
we consider only a low-dimensional hyperparameter vector $(\lambda,\alpha)$,
and use cross-validation over a grid of hyperparameter values, 
solving an inequality constrained PLQ optimization problem for each choice of parameters. 
The model selection process is intuitive, and the entire approach is efficiently implementable,  
since the number of arithmetic operations required for the evaluation of each choice of hyperparameters 
is proportional to that of standard RLS approaches.  
Spline kernel modeling with PLQ optimization 
pave the way to for new applications of
robust, \blue{sparse,} and inequality constrained linear system identification. \\
\noindent
{\bf Acknowledgements} The authors are grateful to Dr. Stephen Becker, for help using TFOCS.  
Dr. Aravkin's research has been funded by the WRF Data Science Professorship.

\bibliographystyle{abbrv}
\bibliography{references_sasha}

\end{document}